\documentclass{article}

\usepackage{arxiv}
\usepackage[utf8]{inputenc} 
\usepackage[T1]{fontenc}    
\usepackage{hyperref}       
\usepackage{url}            
\usepackage{booktabs}       
\usepackage{amsfonts}       
\usepackage{nicefrac}       
\usepackage{microtype}      
\usepackage{lipsum}
\usepackage{fancyhdr}       
\usepackage{graphicx}       
\graphicspath{{media/}}     
\usepackage{amsmath}
\usepackage{multirow}
\usepackage{multicol}
\usepackage{bm}
\usepackage{subcaption}

\title{StoxLSTM: A Stochastic Extended Long Short-Term Memory Network for Time Series Forecasting
}

\author{
  Zihao Wang, Yunjie Li, Lingmin Zan, Zheng Gong \\
  School of Information and Electronics \\
  Beijing Institute of Technology \\
  Beijing, China\\
  \texttt{\{wangzihao24, liyunjie, zanlingmin, gongzheng\}@bit.edu.cn} \\
   \And
  Mengtao Zhu \thanks{Corresponding author} \\
  School of Information and Electronics,  Beijing Institute of Technology\\
  Laboratory of Electromagnetic Space Cognition and Intelligent Control \\
  National Key Laboratory of Science and Technology on Space-Born Intelligent Information Processing\\
  Beijing, China\\
  \texttt{zhumengtao@bit.edu.cn} \\
}

\begin{document}
\maketitle

\begin{abstract}
The Extended Long Short-Term Memory (xLSTM) network has demonstrated strong capability in modeling complex long-term dependencies in time series data. Despite its success, the deterministic architecture of xLSTM limits its representational capacity and forecasting performance, especially on challenging real-world time series datasets characterized by inherent uncertainty, stochasticity, and complex hierarchical latent dynamics. In this work, we propose \textbf{StoxLSTM}, a \textbf{sto}chastic \textbf{xLSTM} within a designed state space modeling framework, which integrates latent stochastic variables directly into the recurrent units to effectively model deep latent temporal dynamics and uncertainty. The designed state space model follows an efficient non-autoregressive generative approach, achieving strong predictive performance without complex modifications to the original xLSTM architecture. Extensive experiments on publicly available benchmark datasets demonstrate that StoxLSTM consistently outperforms state-of-the-art baselines, achieving superior performance and generalization.
\end{abstract}


\section{Introduction}
\label{section1}
Time series forecasting plays a crucial role across diverse domains such as finance \cite{salinas_deepar_2020, Aitian_MMFNet_2024}, energy \cite{zhang_Solar_2023, Uremović_New_2023}, weather \cite{Bi_Nowcasting_2023}, and healthcare \cite{vararlakshmi_parkinson_2021}. Recurrent neural networks (RNNs), particularly Long Short-Term Memory (LSTM) networks \cite{hoch_lstm_1997}, have been widely used due to their capacity to model sequential data \cite{chung_empirical_2014, rangapuram_deep_2018, chang_dilated_2017,wang_pcn_2026}. The Extended Long Short-Term Memory network (xLSTM) \cite{beck_xlstm_2024} further enhances conventional LSTM by introducing exponential gating mechanisms and modified memory structures, enabling better modeling of complex temporal dependencies and improved stability over long sequences. Consequently, xLSTM has demonstrated superior performance across various time series forecasting tasks compared to traditional recurrent models \cite{kong_unlocking_2024, kraus_xlstm-mixer_2024, alharthi_xlstmtime_2024}.

Despite these advances, prior work has predominantly focused on architectural modifications at the top level of xLSTM, without fundamentally redesigning its recurrent units to better capture the intrinsic characteristics of time series data \cite{alharthi_xlstmtime_2024, kraus_xlstm-mixer_2024, kong_unlocking_2024}. Real-world time series often exhibit deep latent dependencies \cite{durbin_time_2012, girin_Dynamical_2021} as well as inherent uncertainty and stochasticity \cite{salinas_deepar_2020, dou_multivariate_2025, chen_probabilistic_2020}, which call for explicit modeling. However, existing xLSTM-based models lack explicit modeling of deep hierarchical latent states, which restricts their capacity to fully represent complex temporal evolution patterns. Moreover, these models are entirely discriminative and deterministic, limiting their ability to model the inherent uncertainty and latent stochastic dynamics commonly present in real-world time series.

\begin{figure}[t]
\centering
\includegraphics[width=1\columnwidth, trim=12 20 12 18, clip]{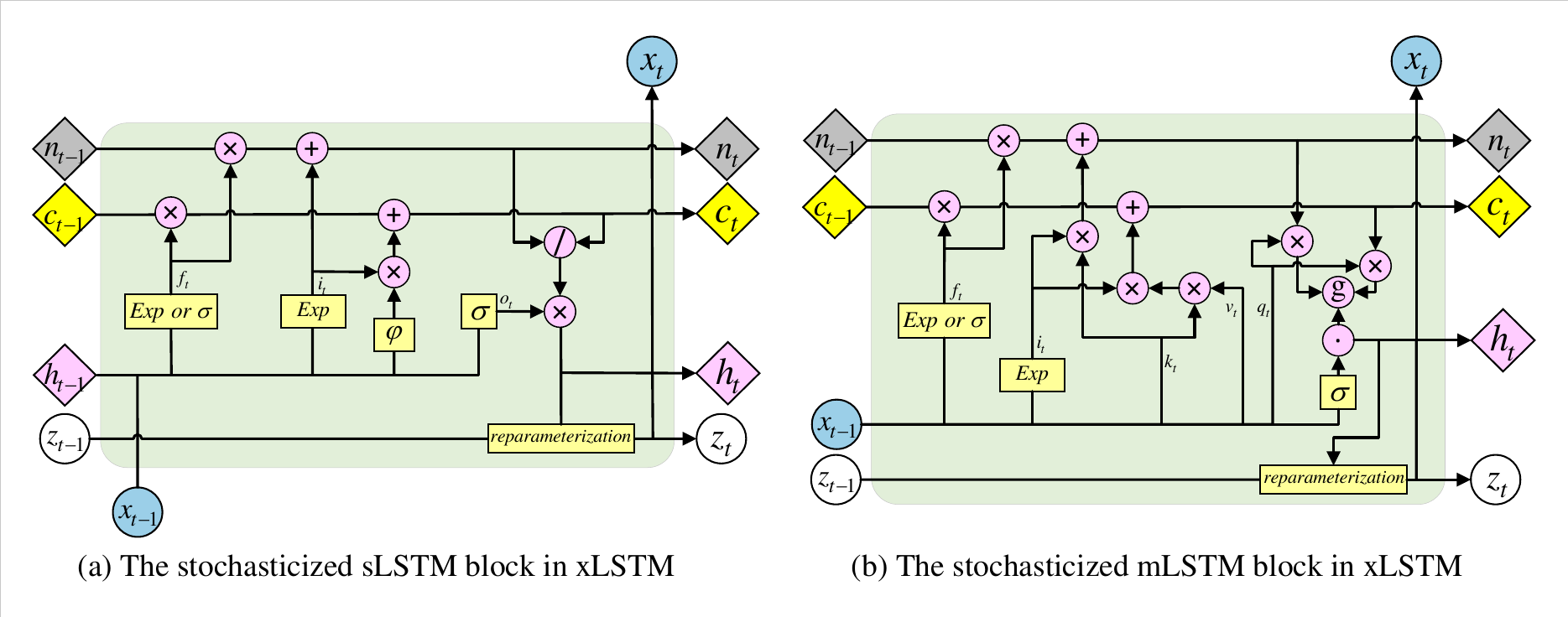} 
\caption{The recurrent unit of StoxLSTM. StoxLSTM integrates stochastic latent variables based on the original xLSTM to represent more complex hierarchical and stochastic characteristics in time series data. The xLSTM consists of two sub-blocks: sLSTM and mLSTM. (a) shows the stochasticized sLSTM block, referred to as StosLSTM, while (b) shows the stochasticized mLSTM block, called StomLSTM. In the figure, diamond shapes denote deterministic variables, such as \(\bm{n}_t\), \(\bm{c}_t\), and \(\bm{h}_t\) from the original xLSTM. Circles indicate stochastic variables, including the latent variable \(\bm{z}_t\) obtained through the reparameterization module, as well as the observed time series variable \(x_t\).}
\label{StoxLSTM_cell}
\end{figure}

To address these limitations, we propose \textbf{StoxLSTM}, a \textbf{sto}chastic \textbf{xLSTM} within a state space modeling framework. First, we mathematically design an efficient, non-autoregressive state space model (SSM) \cite{durbin_time_2012} dependency tailored to the characteristics of the xLSTM recurrent units, which accurately model the temporal evolution of latent variables. Moreover, the SSM dependency facilitates precise and efficient forecasting without introducing excessive complexity to the original xLSTM architecture. Second, we directly integrate latent variables into the xLSTM recurrent units, specifically embedding stochastic latent variables within the sLSTM and mLSTM sub-blocks to form the two stochastic recurrent units—StosLSTM and StomLSTM, as illustrated in Fig. \ref{StoxLSTM_cell}. Based on the designed SSM dependency, the generative and inference models of StoxLSTM are then constructed by leveraging these stochastic recurrent units. Through variational inference, the model effectively captures meaningful latent variables, allowing it to model deep temporal dynamics and uncertainty.

Extensive experiments on multiple publicly available benchmark datasets, such as Electricity, Weather, ETT, Solar, Traffic, PEMS, and ILI \cite{ye_cvacl_2016,tan_from_2015}, demonstrate that StoxLSTM consistently outperforms state-of-the-art baselines in terms of forecasting performance and generalization. Our main contributions are summarized as follows:

\begin{itemize}
    \item A novel, efficient non-autoregressive SSM dependency is designed, tailored to the structural characteristics of xLSTM recurrent units, enabling effective latent variable modeling and enhanced forecasting performance without introducing excessive complexity to the original architecture.
    \item To the best of our knowledge, we propose \textbf{StoxLSTM}, the first \textbf{sto}chastic \textbf{xLSTM} explicitly modeling uncertainty and complex latent dynamics in time series. Its recurrent units integrate latent stochastic variables directly into xLSTM’s sub-blocks. Following the designed SSM dependency, the generative and inference models are formulated based on these stochastic recurrent units and trained via variational inference.
    \item Extensive experiments on diverse publicly available benchmark datasets demonstrate that StoxLSTM consistently outperforms state-of-the-art methods, validating the effectiveness and superiority of the stochastic state space reformulation of xLSTM.
\end{itemize}

\section{Related Work}
\subsection{Time Series Forecasting Models}
Time series forecasting aims to predict future values based on historical observations and has witnessed significant progress through diverse modeling paradigms. Among these, Transformer-based models \cite{vaswani_attention_2017} have attracted extensive attention due to their powerful sequence modeling capabilities. Notable variants include Informer \cite{zhou_informer_2021}, Autoformer \cite{wu_autoformer_2021}, Crossformer \cite{zhang_crossformer_2022}, and FEDformer \cite{zhou_fedformer_2022}, which improve efficiency and accuracy for long-term TSF. However, recent work such as LTSF-Linear \cite{zeng_are_2023} challenges the supremacy of Transformers by demonstrating that simple linear models can outperform them on long-term TSF. These linear models \cite{zeng_are_2023, wang_timemixer_2024, wang_timemixerpuls_2024} better preserve temporal order, effectively utilize longer look-back windows without overfitting, and benefit from enhanced embedding strategies, which also improve interpretability \cite{li_revisiting_2023}. Inspired by these advances, models such as iTransformer \cite{liu_itransformer_2024} and PatchTST \cite{nie_time_2023} integrate linear transformations along the temporal dimension to further boost Transformer-based forecasting performance.

In parallel, recurrent architectures remain competitive. xLSTM \cite{beck_xlstm_2024} has recently gained prominence due to its superior ability to capture complex temporal dependencies and causal relationships inherent in time series data. Compared to Transformer-based models, xLSTM-based models exhibit stronger capabilities in modeling temporal causality, which is critical for accurate forecasting. Moreover, relative to linear models, xLSTM-based models can effectively capture more intricate nonlinear dependencies within time series. Variants such as xLSTMTime \cite{alharthi_xlstmtime_2024}, P-sLSTM \cite{kong_unlocking_2024}, and xLSTM-Mixer \cite{kraus_xlstm-mixer_2024} have demonstrated superior TSF performance compared to both Transformer- and linear-based models, highlighting the effectiveness of xLSTM in modeling long-range temporal patterns.

\subsection{Generative Modeling for Time Series}
Generative modeling has become an important approach for capturing uncertainty and complex temporal dynamics beyond deterministic predictions. These methods introduce latent variables or probabilistic frameworks to model the stochastic nature of real-world time series. Classical statistical models such as ARIMA \cite{box_recent_1968}, VAR \cite{helmut_new_2005}, and SSMs \cite{durbin_time_2012} have long been used for probabilistic forecasting by modeling trends, seasonality, and inter-variable dependencies. However, their strong assumptions and limited capacity to handle nonlinearities and high-dimensional data restrict their applicability in complex real-world scenarios.

Recent advances in deep generative modeling have introduced more flexible frameworks that can better capture complex data distributions without restrictive parametric assumptions. Generative adversarial networks (GANs) \cite{goodfellow_generative_2020, yoon_time-series_2019} and diffusion-based models \cite{ho_denoising_2020, rasul_autoregressive_2021} have demonstrated strong capabilities in modeling intricate temporal patterns and generating realistic synthetic sequences. Among these, deep state space models (SSMs) \cite{girin_Dynamical_2021, rangapuram_deep_2018, tang_probabilistic_2021} stand out by explicitly incorporating latent variables to represent hidden states that evolve over time. This latent variable framework enables deep SSMs to effectively capture complex temporal dependencies that are often nonlinear and stochastic in nature. By modeling the underlying hidden dynamics, deep SSMs can naturally account for the inherent randomness and uncertainty present in real-world time series data, which traditional deterministic models may fail to represent adequately. These deep SSMs typically leverage variational inference techniques \cite{Kingma2013AutoEncodingVB} to approximate intractable posterior distributions, enabling efficient learning and inference. For instance, ProTran \cite{tang_probabilistic_2021} combines SSMs with Transformer architectures to model long-term dependencies in a probabilistic manner, achieving strong performance in uncertainty-aware forecasting tasks.

\section{Methodology}
\subsection{Modeling Time series Forecasting with a Deep SSM}
Consider a multivariate time series \( \bm{x}_{1:L+T}^C \), where \(C\) denotes the number of variables (dimensions), \(L\) denotes the look-back length, and \(T\) denotes the prediction horizon. Each dimension \( x_{1:L+T}^i \) can be treated as an individual univariate time series.

For the univariate time series \( x_{1:L+T} = \{x_1, x_2, \ldots, x_L, \ldots, x_{L+T}\} \), forecasting with deep learning models involves parameterizing the predictive distribution \( p_{\bm\theta}(x_{L+1:L+T} | x_{1:L}) \) with network parameters \(\bm{\theta}\). When integrating deep learning with SSMs, latent states \( \bm{z}_{1:L+T} \) are introduced to capture latent dynamics, transforming the predictive distribution into the joint distribution \( {p_{\bm\theta} }\left( {{x_{1:L + T}},{\bm{z}_{1:L + T}}|{x_{1:L}}} \right) \).

\subsubsection{Dependency Mapping of the Designed SSM} 
\begin{figure}[t]
\centering
\includegraphics[width=0.9\columnwidth, trim=15 15 15 15, clip]{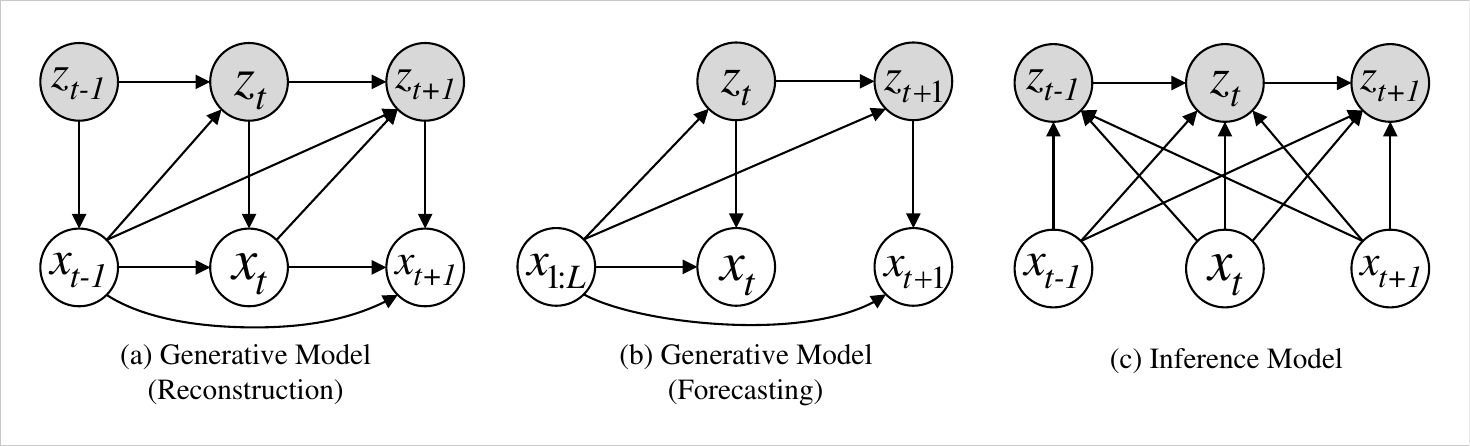} 
\caption{State space models corresponding to the generative model and the inference models in StoxLSTM. (a) and (b) illustrate the reconstruction and forecasting phase of the generative model, respectively. In (a), the state space model depicts the observation \(x_t\) following the conditional distribution \(p_{\bm\theta}(x_t \mid \bm{z}_t, x_{1:t-1})\), while the latent state \(\bm{z}_t\) evolves according to \(p_{\bm\theta}(\bm{z}_t\mid\bm{z}_{t-1}, x_{1:t-1})\). (b) represents the forecasting phase, where \(x_t\) is generated from \(p_{\bm\theta}(x_t\mid \bm{z}_t, x_{1:L})\), and the latent state \(\bm{z}_t\) transitions according to \(p_{\bm\theta}(\bm{z}_t \mid \bm{z}_{t-1}, x_{1:L})\). (c) illustrates the inference model, in which the latent state \(\bm{z}_t\) is inferred from the approximate posterior distribution \(q_\phi(\bm{z}_t \mid \bm{z}_{t-1}, x_{1:L+T})\).}
\label{SSM}
\end{figure}

Considering that the hidden state of xLSTM at each time step already encodes comprehensive information from all past observations \cite{beck_xlstm_2024}, and to avoid introducing excessive complexity into the StoxLSTM recurrent units, we design the generative and inference models in StoxLSTM based on a stochastic SSM framework with explicit dependency assumptions. This design also effectively addresses the error accumulation commonly observed in multi-step autoregressive forecasting \cite{das_decoder-only_2024,zeng_are_2023}. The overall architecture is illustrated in Fig. \ref{SSM}.

Specifically, as shown in Fig. \ref{SSM}(a), the current latent variable \( \bm{z_t} \) is designed to capture all historical observations \( x_{1:t-1} \), while the evolution of latent variables follows the Markov assumption. The current observation \( x_t \) depends exclusively on the current latent state \( \bm{z_t} \) and the entire history of observations \( x_{1:t-1} \). Notably, the SSM dependency structure differs between the reconstruction and forecasting phases: during forecasting, as shown in Fig. \ref{SSM}(b), the model conditions solely on the known observations \( x_{1:L} \), independent of previously predicted outputs. This non-autoregressive formulation effectively mitigates error propagation in multi-step forecasting. Formally, the joint distribution factorizes as:
\begin{equation}
    \begin{aligned}
\begin{array}{l}
\quad {p_{\bm\theta} }\left( {{x_{1:L + T}},{\bm{z}_{1:L + T}}|{x_{1:{\rm{L}}}}} \right) \\
= \prod\limits_{t = 1}^{L + T} {{p_{\bm\theta} }\left( {{x_t}|{x_{1:t - 1}},{\bm{z}_{1:t}},{x_{1:L}}} \right) \cdot {p_{\bm\theta} }\left( {{\bm{z}_t}|{x_{1:t - 1}},{\bm{z}_{1:t - 1}},{x_{1:L}}} \right)} \\
= \prod\limits_{t = 1}^L {{p_{\bm\theta} }\left( {{x_t}|{\bm{z}_t},{x_{1:t - 1}}} \right) \cdot {p_{\bm\theta} }\left( {{\bm{z}_t}|{\bm{z}_{t - 1}},{x_{1:t - 1}}} \right)} \cdot \prod\limits_{t = L + 1}^{L + T} {{p_{\bm\theta} }\left( {{x_t}|{\bm{z}_t},{x_{1:L}}} \right) \cdot {p_{\bm\theta} }\left( {{\bm{z}_t}|{\bm{z}_{t - 1}},{x_{1:L}}} \right)} 
\end{array}\label{eq1}
    \end{aligned}
\end{equation}
wherein $\prod\limits_{t = 1}^L {{p_{\bm\theta} }\left( {{x_t}|{\bm{z}_t},{x_{1:t - 1}}} \right) \cdot {p_{\bm\theta} }\left( {{\bm{z}_t}|{\bm{z}_{t - 1}},{x_{1:t - 1}}} \right)}$ corresponds to the reconstruction of the historical observations \( x_{1:t-1} \). While $\prod\limits_{t = L + 1}^{L + T} {{p_{\bm\theta} }\left( {{x_t}|{\bm{z}_t},{x_{1:L}}} \right) \cdot {p_{\bm\theta} }\left( {{\bm{z}_t}|{\bm{z}_{t - 1}},{x_{1:L}}} \right)}$ models the evolution of future latent states and observations conditioned on the known observations \( x_{1:L} \).

The inference model estimates the approximate posterior distribution of the latent variables. As illustrated in Fig. \ref{SSM}(c), the current latent variable \( \bm{z_t} \) in the inference model has access to the entire observation \( x_{1:L+T} \). The evolution of latent variables similarly follows the Markov assumption. Accordingly, the approximate posterior over the latent states \( \bm{z}_{1:L+T} \), parameterized by the inference model parameters \(\bm{\varphi}\), factorizes as:
\begin{equation}
\begin{aligned}
    {q_{\bm\varphi} }\left( {{\bm{z}_{1:L + T}}|{x_{1:L + T}}} \right) = \prod\limits_{t = 1}^{L + T} {{q_{\bm\varphi}}\left( {{\bm{z}_t}|{\bm{z}_{1:t - 1}},{x_{1:L + T}}} \right)}
= \prod\limits_{t = 1}^{L + T} {{q_{\bm\varphi} }\left( {{\bm{z}_t}|{\bm{z}_{t - 1}},{x_{1:L + T}}} \right)} \label{eq2}
\end{aligned}
\end{equation}

\subsection{A Stochastic Extended Long Short-term Memory Network}

\begin{figure}
    \centering
    \includegraphics[width=1\textwidth]{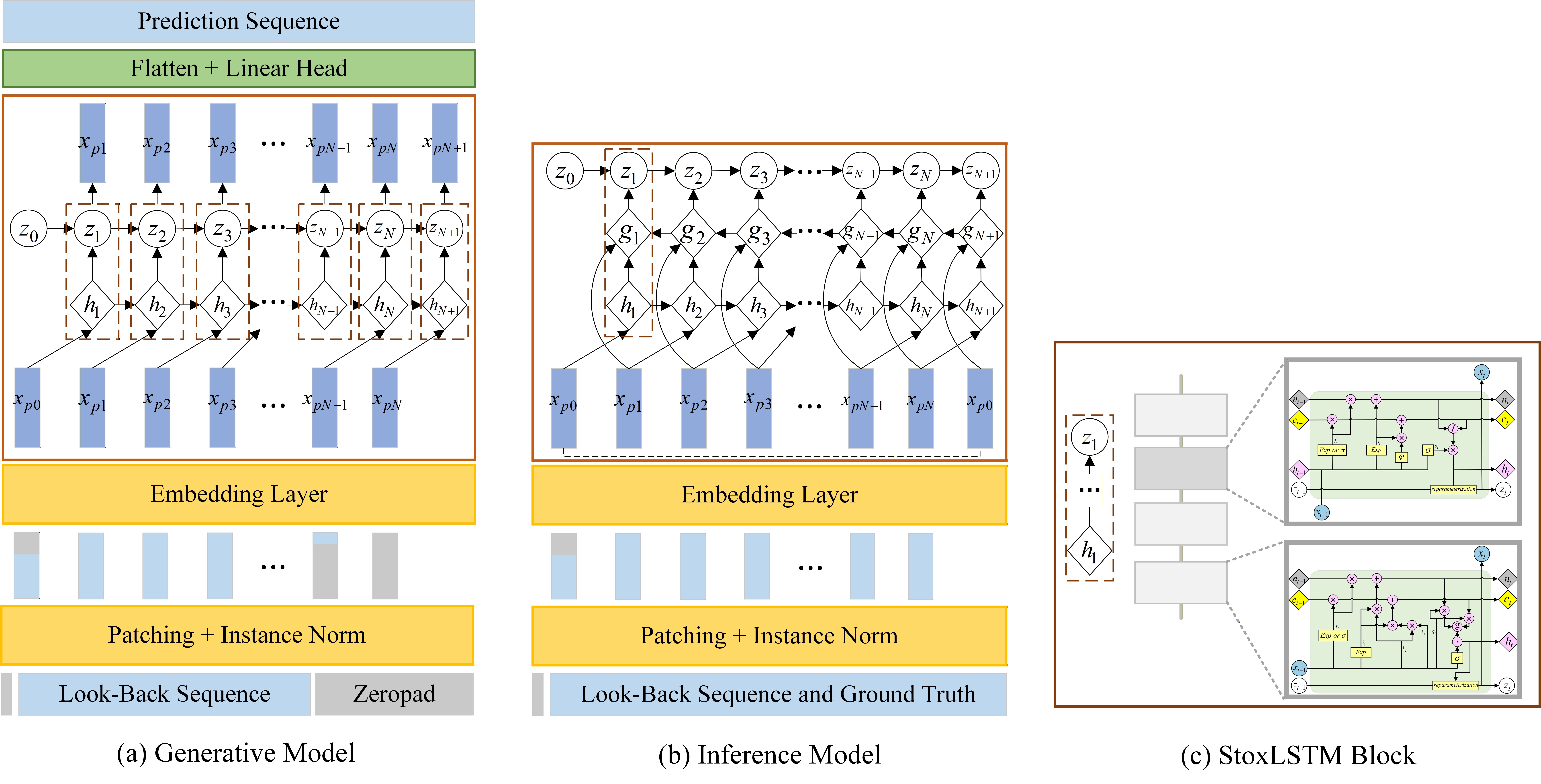}
    \caption{Overall framework of StoxLSTM. (a) depicts the generative model, while (b) illustrates the inference model. In both (a) and (b), the dashed boxes represent the stacked recurrent units of StoxLSTM shown in (c). Each stacked unit in (c) corresponds to either a StomLSTM or a StosLSTM block.}
    \label{StoxLSTM_Backbone}
\end{figure}

The overall structure of StoxLSTM is illustrated in Fig. \ref{StoxLSTM_Backbone}. StoxLSTM performs channel-independent forecasting for multivariate time series by dividing the multivariate time series into multiple individual univariate sequences.  Each univariate sequence \( \bm{x_i} \in \mathbb{R}^{1 \times L}, i=1,2,\ldots,C \) is processed independently by the StoxLSTM module, and the outputs are concatenated to form the multivariate prediction \( \bm{\widehat{x}} \in \mathbb{R}^{C \times T} \). To address the challenges of modeling long sequences encountered by xLSTM \cite{kong_unlocking_2024}, each univariate sequence is further segmented into consecutive patches. Section \ref{Recurrent Unit in StoxLSTM} introduces the two recurrent units of StoxLSTM, namely StosLSTM and StomLSTM, while Sections \ref{generative model} and \ref{inference model} describe how the generative and inference models of StoxLSTM are constructed based on these recurrent units.

\subsubsection{Design of the Recurrent Unit in StoxLSTM} 
\label{Recurrent Unit in StoxLSTM}
By incorporating stochastic latent variables \( \bm{z}_{1:t} \) into the original xLSTM recurrent unit, StoxLSTM follows the SSMs that respect temporal causality, Markovianity, and non-autoregressive dynamics, as shown in Figs. \ref{StoxLSTM_cell} and \ref{SSM}. Specifically, at each time step \( t \), the latent state \( \bm{z}_t \) is modeled as a stochastic variable conditioned on the previous latent state \( \bm{z}_{t-1} \) and the current deterministic hidden state \( \bm{h}_t \).

The forward pass of StosLSTM (the stochasticized sLSTM block in xLSTM, as shown in Fig. \ref{StoxLSTM_cell}(a)) is as follows:

\[
\begin{array}{ll}
{\bm{c}_t} = {\bm{f}_t}^\prime {\bm{c}_{t - 1}} + {\bm{i}_t}^\prime {\bm{y}_t} & \text{cell state} \\
{\bm{n}_t} = {\bm{f}_t}^\prime {\bm{n}_{t - 1}} + {\bm{i}_t}^\prime & \text{normalizer state} \\
{\bm{h}_t} = {\bm{o}_t} {\bm{\widetilde{h}}_t}, \quad {\bm{\widetilde{h}}_t} = {\bm{c}_t}/{\bm{n}_t} & \text{hidden state} \\
\bm{z}_t \sim \mathcal{N}\big( \mu_{\bm\theta}(\bm{h}_t, \bm{z}_{t-1}), \sigma_{\bm\theta}(\bm{h}_t, \bm{z}_{t-1}) \big) & \text{latent state} \\
{x_t} = \varphi \left( {\bm{w_z}^T{\bm{z}_t} + \bm{r_z}{\bm{h}_t} + \bm{b_z}} \right) & \text{output generation}
\end{array}
\]
Here, \( \bm{y}_t \) denotes the cell input. The variables \( \bm{c}_t \), \( \bm{n}_t \), and \( \bm{h}_t \) represent the cell state, normalizer state, and hidden state, respectively. The latent state (latent variable) \( \bm{z}_t \) follows a Gaussian distribution parameterized by mean \( \mu_{\bm\theta}(\cdot) \) and standard deviation \( \sigma_{\bm\theta}(\cdot) \), both implemented as fully connected neural networks. The terms \( {\bm{f}_t}^\prime \) and \( {\bm{i}_t}^\prime \) correspond to the stabilizer’s forget and input gates \cite{beck_xlstm_2024}, while \( \bm{o}_t \) denotes the output gate. Parameters \( \bm{w_z} \), \( \bm{r_z} \), and \( \bm{b_z} \) are the weights and bias associated with \( \bm{z}_t \) and \( \bm{h}_t \). The function \( \varphi \) represents the activation function. Finally, \( x_t \) denotes the generated output (observation) at time step \( t \).

Similarly, the forward pass of StomLSTM (the stochasticized mLSTM block in xLSTM, as shown in Fig. \ref{StoxLSTM_cell}(b)) is:

\[
\begin{array}{ll}
{\bm{c}_t} = {\bm{f}_t}{\bm{c}_{t - 1}} + {\bm{i}_t}{\bm{v}_t}\bm{k}_t^T & \text{cell state} \\
{\bm{n}_t} = {\bm{f}_t}{\bm{n}_{t - 1}} + {\bm{i}_t}{\bm{k}_t} & \text{normalizer state} \\
{\bm{h}_t} = {\bm{o}_t} \odot \bm{\widetilde h_t}, \quad \bm{\widetilde h_t} = \bm{C_t}{\bm{q}_t}/max\left\{ {\left| {\bm{n}_t^T{\bm{q}_t}} \right|,1} \right\} & \text{hidden state} \\
\bm{z}_t \sim \mathcal{N}\big( \mu_{\bm\theta}(\bm{h}_t, \bm{z}_{t-1}), \sigma_{\bm\theta}(\bm{h}_t, \bm{z}_{t-1}) \big) & \text{latent state} \\
{x_t} = \varphi \left( {\bm{w_z}^T{z_t} + \bm{r_z}{\bm{h}_t} + \bm{b_z}} \right) & \text{output generation}
\end{array}
\]
Here, \(\bm{k}_t\) and \(\bm{q}_t\) denote the key and query inputs, respectively, and \(\odot\) represents element-wise multiplication.

In both StosLSTM and StomLSTM, the latent state \(\bm{z}_t\) is modeled as a Gaussian random variable conditioned on \(\bm{h}_t\) and \(\bm{z}_{t-1}\). This probabilistic formulation enables the model to capture inherent uncertainty and stochasticity in real-world time series. By sampling \(\bm{z}_t\) via the reparameterization block, StoxLSTM can represent diverse possible future trajectories, effectively modeling the variability and noise intrinsic to temporal dynamics. Consequently, the latent state update follows the conditional distribution \( p(\bm{z}_t \mid \bm{z}_{t-1}, x_{1:t-1}) \), while the output \( x_t \) is generated according to \( p(x_t \mid \bm{z}_t, x_{1:t-1}) \).

\subsubsection{Generative Model}
\label{generative model}
The generative process adheres to the factorization presented in Eq. \eqref{eq1}. During preprocessing, zero-padding is applied at both ends of the look-back sequence: padding at the front generates the initial input \( x_0 \) with a length equal to the patch stride \( S \), and padding at the end aligns the predicted sequence length \( T \).  The padded sequence is then segmented into patches \( \bm{p}_{0:N} \), where \( N = \lceil \frac{L + T + S - P}{S} \rceil \) and \( P \) denotes the patch size. These patches are embedded as follows:

\begin{equation}
{\bm{x}_{p0:pN}} = {\rm{Embedding}}({\rm{patching}}([\underbrace {0, \ldots ,0,}_{{S}}{x_{1:L}},\underbrace {0, \ldots ,0}_T])) \label{eq4}
\end{equation}

The embedded patches $\bm{x}_{p0:pN}$ are fed into the StoxLSTM recurrent unit to generate latent states \( \bm{z}_{1:N+1} \) and time series patches \(\bm{x}_{p1:pN+1}\):
\begin{equation}
    \bm{z}_{1:N+1}, \bm{x}_{p1:pN+1} = \mathrm{RecurrentUnit}(\bm{x}_{p0:pN}) \label{eq6}
\end{equation}

Therefore, during the reconstruction phase, the latent state \( \bm{z}_t \) and observation \( x_t \) are generated from the probability distributions \( {p_{\bm\theta}}(\bm{z}_t | \bm{z}_{t-1}, x_{1:t-1}) \) and \( {p_{\bm\theta}}(x_t | \bm{z}_t, x_{1:t-1}) \), respectively. While during the forecasting phase, \( \bm{z}_t \) and \( x_t \) are generated from the distributions \( {p_{\bm\theta}}(\bm{z}_t | \bm{z}_{t-1}, x_{1:L}) \) and \( {p_{\bm\theta}}(x_t | \bm{z}_t, x_{1:L}) \), respectively.

Finally, The time series patches \( \bm{x}_{p1:pN+1} \) are flattened and passed through a fully connected layer to generate the final prediction \( x_{L+1:L+T} \):
\begin{equation}
    x_{L+1:L+T} = \mathrm{FCL}(\mathrm{Flatten}(\bm{x}_{p1:pN+1})) \label{eq7}
\end{equation}
where \(\mathrm{FCL}(\cdot)\) denotes a fully connected layer.

\subsubsection{Inference Model} \label{inference model}
The inference model follows the factorization in Eq. \eqref{eq2} and employs a bidirectional recurrent unit \cite{Kingma2013AutoEncodingVB,girin_Dynamical_2021}. Zero-padding is applied at the beginning of the sequence \( x_{1:L+T} \) with length equal to the patch stride \( S \), followed by patching and embedding:
\begin{equation}
{\bm{x}_{p0:pN}} = {\rm{Embedding}}({\rm{patching}}([\underbrace {0, \ldots ,0,}_{\rm{S}}{x_{1:L + T}}])) \label{eq8}
\end{equation}

The bidirectional recurrent unit processes these embedded patches, producing forward and backward hidden states \(\bm{h}_{1:N + 1}\) and \( \bm{g}_{N+1:1} \), which are then used to infer the latent states \( \bm{z}_{1:N+1} \):
\begin{equation}
    {\bm{z}_{1:N + 1}}, \_\ = \mathrm{RecurrentUnit_{bidirection}}(\bm{x}_{p0:pN})  \label{eq9}
\end{equation}

The bidirectional recurrent unit enables the latent states \( \bm{z}_t \) to be drawn from the approximate posterior distribution \({q_{\bm\varphi} }\left( {{\bm{z}_t}|{\bm{z}_{t - 1}},{x_{1:L + T}}} \right)\). The role of the inference model is to obtain this approximate posterior during training, while during testing, only the generative model is utilized.

\subsubsection{Evidence Lower Bound of StoxLSTM}
Following the variational inference framework \cite{girin_Dynamical_2021, tang_probabilistic_2021, bayer_learning_2015, fraccaro_sequential_2016, wang_mspf_2025}, we derive the Evidence Lower Bound (ELBO) as the training objective for StoxLSTM based on the factorization assumptions in Eqs. \eqref{eq1} and \eqref{eq2}:
\[\begin{array}{*{20}{l}}
{\log p\left( {{x_{1:L + T}}|{x_{1:L}}} \right) \ge L\left( {\theta ,\phi ;{x_{1:L + T}}} \right) = }\\
\begin{array}{l}
\quad \quad \sum\limits_{t = 1}^L {\left( {{E_{{q_{\bm\varphi} }\left( {{\bm{z}_t}|{\bm{z}_{t - 1}},{x_{1:L + T}}} \right)}}\left[ {\log {p_{\bm\theta} }\left( {{x_t}|{\bm{z}_t},{x_{1:t - 1}}} \right)} \right]} \right.}
 - \left. {KL\left( {{q_{\bm\varphi} }\left( {{\bm{z}_t}|{\bm{z}_{t - 1}},{x_{1:L + T}}} \right)||{p_{\bm\theta} }\left( {{\bm{z}_t}|{\bm{z}_{t - 1}},{x_{1:t - 1}}} \right)} \right)} \right)
\end{array}\\
\begin{array}{l}
\qquad  + \sum\limits_{t = L + 1}^{L + T} {\left( {{E_{{q_{\bm\varphi} }\left( {{\bm{z}_t}|{\bm{z}_{t - 1}},{x_{1:L + T}}} \right)}}\left[ {\log {p_{\bm\theta} }\left( {{x_t}|{\bm{z}_t},{x_{1:L}}} \right)} \right]} \right.} 
 \left. { - KL\left( {{q_{\bm\varphi} }\left( {{\bm{z}_t}|{\bm{z}_{t - 1}},{x_{1:L + T}}} \right)||{p_{\bm\theta} }\left( {{\bm{z}_t}|{\bm{z}_{t - 1}},{x_{1:L}}} \right)} \right)} \right)
\end{array}
\end{array}\]
Here, $KL(\cdot)$ denotes the KL divergence. This objective encourages the model to accurately reconstruct the observed sequence while maintaining consistency between the inferred latent states and their temporal dynamics. Detailed derivations and implementation details are provided in \ref{app A}.


\section{Experiments} \label{experiments}
\subsection{Time Series Forecasting}

\subsubsection{Datasets}
We employ 9 publicly available datasets to evaluate the performance of our proposed StoxLSTM for long-term time series forecasting. These datasets include: Electricity\footnote{\url{https://archive.ics.uci.edu/ml/datasets/ElectricityLoadDiagrams20112014}}, Weather\footnote{\url{https://www.bgc-jena.mpg.de/wetter/}}, Solar\footnote{\url{https://github.com/laiguokun/multivariate-time-series-data}}, Traffic\footnote{\url{http://pems.dot.ca.gov}}, ILI\footnote{\url{https://gis.cdc.gov/grasp/fluview/fluportaldashboard.html}}, and four ETT\footnote{\url{https://github.com/zhouhaoyi/ETDataset}} datasets (ETTm1, ETTm2, ETTh1, ETTh2). Additionally, we utilize 4 public datasets to assess the performance of short-term time series forecasting, which include: PEMS\footnote{\url{http://pems.dot.ca.gov}} (PEMS03, PEMS04, PEMS07, and PEMS08). These datasets have been widely used for benchmarking and are publicly available in \cite{wu_autoformer_2021}, \cite{zeng_are_2023}, \cite{kraus_xlstm-mixer_2024}, and \cite{lai_modeling_2018}. Detailed dataset information is summarized in Table \ref{dataset}.

\begin{table}[!htbp]
    \caption{Dataset detailed descriptions. The dataset size is organized in (Train, Validation, Test).}
    \label{dataset}
    \centering
    \scriptsize
    \begin{tabular}{l|c|c|c|c|c|c}
        \toprule
        Datasets & Dim & Horizon & Frequency & Dataset size & Information & Tasks \\
        \midrule
        Weather & 21 & \{96, 192, 336, 720\} & 10 min & (36887,5270,10539) & Weather & Long-term forecasting \\
        \midrule
        Electricity & 321 & \{96, 192, 336, 720\} & Hourly & (18412,2632,5260) & Electricity & Long-term forecasting \\
        \midrule
        Solar-Energy & 137 & \{96, 192, 336, 720\} & 10 min & (36792,5256,10512) & Electricity & Long-term forecasting \\
        \midrule
        ETTh1, ETTh2 & 7 & \{96, 192, 336, 720\} & Hourly & (8640,2880,2880) & Temperature & Long-term forecasting \\
        \midrule
        ETTm1, ETTm2 & 7 & \{96, 192, 336, 720\} & 15 min & (34560,11520,11520) & Temperature & Long-term forecasting \\
        \midrule
        Traffic & 862 & \{96, 192, 336, 720\} & Hourly & (12280,1756,3508) & Transportation & Long-term forecasting  \\
        \midrule
        ILI & 170 & \{24, 36, 48, 60\} & 7 days & (676,97,193) & Illness & Long-term forecasting \\
        \midrule
        PEMS03 & 358 & 12 & 5 min & (15724,5241,5243) & Transportation & Short-term forecasting \\
        \midrule
        PEMS04 & 307 & 12 & 5 min & (10195,3398,3399) & Transportation & Short-term forecasting \\
        \midrule
        PEMS07 & 883 & 12 & 5 min & (16934,5644,5646) & Transportation & Short-term forecasting \\
        \midrule
        PEMS08 & 170 & 12 & 5 min & (10713,3571,3572) & Transportation & Short-term forecasting \\
        \bottomrule
    \end{tabular}
\end{table}

\subsubsection{Baselines and Experiment Settings.}
We select three types of state-of-the-art models as baselines for long-term and short-term time series forecasting:

\textbf{Transformer-based}: Informer \cite{zhou_informer_2021}, Autoformer \cite{wu_autoformer_2021}, Crossformer \cite{zhang_crossformer_2022}, FEDformer \cite{zhou_fedformer_2022}, PatchTST \cite{nie_time_2023}, and iTransformer \cite{liu_itransformer_2024}.

\textbf{Linear-based}: Dlinear \cite{zeng_are_2023} and TimeMixer++ \cite{wang_timemixer_2024, wang_timemixerpuls_2024}.

\textbf{xLSTM-based}: xLSTMTime \cite{alharthi_xlstmtime_2024}, P-sLSTM \cite{kong_unlocking_2024}, and xLSTM-Mixer \cite{kraus_xlstm-mixer_2024}.

All models are evaluated under consistent experimental conditions.

For long-term forecasting, the prediction horizons for the ILI dataset are set to \( T \in \{24, 36, 48, 60\} \), with StoxLSTM’s look-back window fixed at \( L = 96 \). For all other datasets, the prediction horizons are \( T \in \{96, 192, 336, 720\} \), and the look-back length of StoxLSTM is set to \( L = 336 \). For short-term forecasting, all models are set to \( T = 12 \) and \( L = 96 \). Baseline results and look-back length \(L\) settings for long-term forecasting are primarily sourced from PatchTST \cite{nie_time_2023}, xLSTM-Mixer \cite{kraus_xlstm-mixer_2024}, and TimeMixer++ \cite{wang_timemixerpuls_2024}, supplemented by our own experiments where results are unavailable. Evaluation metrics for long-term forecasting include Mean Squared Error (MSE) and Mean Absolute Error (MAE), while short-term forecasting is assessed using MAE, Mean Absolute Percentage Error (MAPE), and Root Mean Squared Error (RMSE). Experiments are conducted on an NVIDIA RTX 3090 GPU (24GB) and 3 NVIDIA A800 GPUs (80GB).

\subsubsection{Results}
Table \ref{tab:1} and Table \ref{tab:2} present the results of long- and short-term time series forecasting, respectively. In general, StoxLSTM consistently outperforms all baseline models in nearly all datasets and prediction horizons, achieving the lowest MSE, MAE, MAPE, and RMSE. 

\begin{table}[!htbp]
    \caption{Long-term time series forecasting results. The best results are in \textbf{ bold}, and the second best are in \underline{ underline}.}
    \label{tab:1}
    \centering
    \setlength{\tabcolsep}{2pt} 
    \setlength{\abovecaptionskip}{0.cm}
    \scriptsize
    \begin{tabular}{c|c|cc|cc|cc|cc|cc|cc|cc|cc|cc|cc|cc}
    \toprule
    \multicolumn{2}{c|}{Models} & \multicolumn{2}{c|}{StoxLSTM} & \multicolumn{2}{c|}{xLSTM-Mixer} & \multicolumn{2}{c|}{xLSTMTime} & \multicolumn{2}{c|}{P-sLSTM} & \multicolumn{2}{c|}{iTransformer} & \multicolumn{2}{c|}{PatchTST} & \multicolumn{2}{c|}{FEDformer} & \multicolumn{2}{c|}{Informer} & \multicolumn{2}{c|}{Autoformer} & \multicolumn{2}{c|}{Dlinear} & \multicolumn{2}{c}{TimeMixer++} \\ \midrule
    \multicolumn{2}{c|}{Metric} & MSE & MAE & MSE & MAE & MSE & MAE & MSE & MAE & MSE & MAE & MSE & MAE & MSE & MAE & MSE & MAE & MSE & MAE & MSE & MAE & MSE & MAE\\ \midrule
    \multirow{4}{*}{\rotatebox{90}{Weather}} 
    & 96 & \textbf{0.111} & \textbf{0.164} & \underline{0.143} & \underline{0.184} & 0.144 & 0.187 & 0.149 & 0.208 & 0.174 & 0.214 & 0.149 & 0.198 & 0.217 & 0.296 & 0.300 & 0.384 & 0.266 & 0.336 & 0.176 & 0.237 & 0.155 & 0.205 \\
    & 192 & \textbf{0.136} & \textbf{0.198} & \underline{0.186} & \underline{0.226} & 0.192 & 0.236 & 0.197 & 0.256 & 0.221 & 0.254 & 0.194 & 0.241 & 0.276 & 0.336 & 0.598 & 0.544 & 0.307 & 0.367 & 0.220 & 0.282 & 0.201 & 0.245 \\
    & 336 & \textbf{0.162} & \textbf{0.227} & \underline{0.236} & \underline{0.266} & 0.237 & 0.272 & 0.249 & 0.297 & 0.278 & 0.296 & 0.245 & 0.282 & 0.339 & 0.380 & 0.578 & 0.523 & 0.359 & 0.395 & 0.265 & 0.319 & 0.237 & 0.265 \\
    & 720 & \textbf{0.195} & \textbf{0.255} & \underline{0.310} & \underline{0.323} & 0.313 & 0.326 & 0.320 & 0.350 & 0.358 & 0.347 & 0.314 & 0.334 & 0.403 & 0.428 & 1.059 & 0.741 & 0.419 & 0.428 & 0.323 & 0.362 & 0.312 & 0.334 \\ \midrule
    \multirow{4}{*}{\rotatebox{90}{Electricity}}
    & 96 & \textbf{0.117} & 0.223 & \underline{0.126} & \textbf{0.218} & 0.128 & \underline{0.221} & 0.130 & 0.226 & 0.148 & 0.240 & 0.129 & 0.222 & 0.193 & 0.308 & 0.274 & 0.368 & 0.201 & 0.317 & 0.140 & 0.237 & 0.135 & 0.222 \\
    &192 & \textbf{0.136} & 0.242 & \underline{0.144} & \textbf{0.235} & 0.150 & 0.243 & 0.148 & 0.243 & 0.162 & 0.253 & 0.147 & \underline{0.240} & 0.201 & 0.315 & 0.296 & 0.386 & 0.222 & 0.334 & 0.153 & 0.249 & 0.147 & \textbf{0.235} \\ 
    & 336 & \textbf{0.144} & 0.253 & \underline{0.157} & \underline{0.250} & 0.166 & 0.259 & 0.165 & 0.262 & 0.178 & 0.269 & 0.163 & 0.259 & 0.214 & 0.329 & 0.300 & 0.394 & 0.231 & 0.338 & 0.169 & 0.267 & 0.164 & \textbf{0.245} \\
    & 720 & \textbf{0.159} & \textbf{0.270} & \underline{0.183} & \underline{0.276} & 0.185 & \underline{0.276} & 0.199 & 0.293 & 0.225 & 0.317 & 0.197 & 0.290 & 0.246 & 0.355 & 0.373 & 0.439 & 0.254 & 0.331 & 0.203 & 0.301 & 0.212 & 0.310 \\ \midrule
    \multirow{4}{*}{\rotatebox{90}{Solar}}
    & 96 & \textbf{0.098} & \textbf{0.179} & 0.227 & 0.313 & 0.241 & 0.306 & \underline{0.167} & 0.232 & 0.203 & 0.237 & \underline{0.167} & \underline{0.224} & 0.287 & 0.383 & 0.200 & 0.247 & 0.456 & 0.446 & 0.289 & 0.377 & 0.171 & 0.231 \\
    & 192 & \textbf{0.124} & \textbf{0.212} & 0.245 & 0.328 & 0.265 & 0.331 & \underline{0.180} & \underline{0.241} & 0.233 & 0.261 & 0.189 & 0.251 & 0.278 & 0.364 & 0.220 & 0.251 & 0.588 & 0.561 & 0.319 & 0.397 & 0.218 & 0.263 \\
    & 336 & \textbf{0.139} & \textbf{0.224} & 0.255 & 0.330 & 0.267 & 0.336 & \underline{0.190} & \underline{0.248} & 0.248 & 0.273 & 0.193 & 0.252 & 0.319 & 0.397 & 0.260 & 0.287 & 0.595 & 0.588 & 0.352 & 0.415 & 0.212 & 0.269 \\
    & 720 & \textbf{0.169} & \textbf{0.241} & 0.257 & 0.324 & 0.261 & 0.324 & \underline{0.196} & \underline{0.249} & 0.249 & 0.275 & 0.204 & 0.269 & 0.319 & 0.402 & 0.244 & 0.301 & 0.733 & 0.633 & 0.356 & 0.412 & 0.212 & 0.270 \\ \midrule
    \multirow{4}{*}{\rotatebox{90}{ETTh1}}
    & 96 & \textbf{0.339} & 0.394 & \underline{0.359} & \textbf{0.386} & 0.368 & 0.395 & 0.381 & 0.405 & 0.386 & 0.405 & 0.370 & 0.400 & 0.376 & 0.419 & 0.865 & 0.713 & 0.449 & 0.459 & 0.375 & \underline{0.399} & 0.361 & 0.403 \\
    & 192 & \textbf{0.370} & \textbf{0.411} & 0.402 & 0.417 & \underline{0.401} & \underline{0.416} & 0.420 & 0.431 & 0.441 & 0.436 & 0.413 & 0.429 & 0.420 & 0.448 & 1.008 & 0.792 & 0.500 & 0.482 & 0.405 & 0.416 & 0.416 & 0.441 \\
    & 336 & \textbf{0.379} & \textbf{0.419} & \underline{0.408} & \underline{0.429} & 0.422 & 0.437 & 0.456 & 0.458 & 0.487 & 0.458 & 0.422 & 0.440 & 0.459 & 0.465 & 1.107 & 0.809 & 0.521 & 0.496 & 0.439 & 0.443 & 0.430 & 0.434 \\
    & 720 & \textbf{0.406} & \textbf{0.443} & \underline{0.419} & \underline{0.448} & 0.441 & 0.465 & 0.516 & 0.512 & 0.503 & 0.491 & 0.447 & 0.468 & 0.506 & 0.507 & 1.181 & 0.865 & 0.514 & 0.512 & 0.472 & 0.490 & 0.467 & 0.451 \\ \midrule
    \multirow{4}{*}{\rotatebox{90}{ETTh2}}
    & 96 & \textbf{0.237} & \textbf{0.307} & \underline{0.267} & \underline{0.329} & 0.273 & 0.333 & 0.320 & 0.378 & 0.297 & 0.349 & 0.274 & 0.337 & 0.346 & 0.388 & 3.755 & 1.525 & 0.358 & 0.397 & 0.289 & 0.353 & 0.276 & 0.328 \\
    & 192 & \textbf{0.264} & \textbf{0.336} & \underline{0.338} & \underline{0.375} & 0.340 & 0.378 & 0.422 & 0.442 & 0.380 & 0.400 & 0.341 & 0.382 & 0.429 & 0.441 & 5.602 & 1.931 & 0.456 & 0.452 & 0.383 & 0.418 & 0.342 & 0.379 \\
    & 336 & \textbf{0.278} & \textbf{0.353} & 0.367 & 0.401 & 0.373 & 0.403 & 0.523 & 0.494 & 0.428 & 0.432 & \underline{0.329} & \underline{0.384} & 0.496 & 0.487 & 4.721 & 1.835 & 0.482 & 0.486 & 0.448 & 0.465 & 0.346 & 0.398 \\
    & 720 & \textbf{0.310} & \textbf{0.380} & \underline{0.388} & 0.424 & 0.398 & 0.430 & 0.946 & 0.669 & 0.427 & 0.445 & 0.379 & 0.422 & 0.463 & 0.474 & 3.647 & 1.625 & 0.515 & 0.511 & 0.605 & 0.551 & 0.392 & \underline{0.415} \\ \midrule
    \multirow{4}{*}{\rotatebox{90}{ETTm1}}
    & 96 & \textbf{0.223} & \textbf{0.310} & \underline{0.275} & \underline{0.328} & 0.286 & 0.335 & 0.292 & 0.343 & 0.334 & 0.368 & 0.293 & 0.346 & 0.379 & 0.419 & 0.672 & 0.571 & 0.505 & 0.475 & 0.299 & 0.343 & 0.310 & 0.340 \\
    & 192 & \textbf{0.261} & \textbf{0.339} & \underline{0.319} & \underline{0.354} & 0.329 & 0.361 & 0.329 & 0.369 & 0.377 & 0.391 & 0.333 & 0.370 & 0.426 & 0.441 & 0.795 & 0.669 & 0.553 & 0.496 & 0.335 & 0.365 & 0.348 & 0.362 \\
    & 336 & \textbf{0.290} & \textbf{0.360} & \underline{0.353} & \underline{0.374} & 0.358 & 0.379 & 0.362 & 0.391 & 0.426 & 0.420 & 0.369 & 0.392 & 0.445 & 0.459 & 1.212 & 0.871 & 0.621 & 0.537 & 0.369 & 0.386 & 0.376 & 0.391 \\
    & 720 & \textbf{0.328} & \textbf{0.385} & \underline{0.409} & \underline{0.407} & 0.416 & 0.411 & 0.421 & 0.424 & 0.491 & 0.459 & 0.416 & 0.420 & 0.543 & 0.490 & 1.166 & 0.823 & 0.671 & 0.561 & 0.425 & 0.421 & 0.440 & 0.423 \\ \midrule
    \multirow{4}{*}{\rotatebox{90}{ETTm2}}
    & 96 & \textbf{0.122} & \textbf{0.220} & \underline{0.157} & \underline{0.244} & 0.164 & 0.250 & 0.238 & 0.302 & 0.180 & 0.264 & 0.166 & 0.256 & 0.203 & 0.287 & 0.365 & 0.453 & 0.255 & 0.339 & 0.167 & 0.260 & 0.170 & 0.245 \\
    & 192 & \textbf{0.160} & \textbf{0.258} & \underline{0.213} & \underline{0.285} & 0.218 & 0288 & 0.337 & 0.376 & 0.250 & 0.309 & 0.223 & 0.296 & 0.269 & 0.328 & 0.533 & 0.563 & 0.281 & 0.340 & 0.224 & 0.303 & 0.229 & 0.291 \\
    & 336 & \textbf{0.189} & \textbf{0.284} & \underline{0.269} & \underline{0.322} & 0.271 & 0.322 & 0.444 & 0.438 & 0.311 & 0.348 & 0.274 & 0329 & 0.325 & 0.366 & 1.363 & 0.887 & 0.339 & 0.372 & 0.281 & 0.342 & 0.303 & 0.343 \\
    & 720 & \textbf{0.248} & \textbf{0.335} & \underline{0.351} & \underline{0.377} & 0.361 & 0.380 & 0.488 & 0.482 & 0.412 & 0.407 & 0.362 & 0.385 & 0.421 & 0.415 & 3.379 & 1.388 & 0.422 & 0.419 & 0.397 & 0.421 & 0.373 & 0.399 \\ \midrule
    \multirow{4}{*}{\rotatebox{90}{Traffic}}
    & 96 & \textbf{0.333} & 0.258 & \underline{0.351} & \textbf{0.236} & 0.358 & \underline{0.242} & 0.576 & 0.332 & 0.395 & 0.268 & 0.360 & 0.249 & 0.587 & 0.366 & 0.719 & 0.391 & 0.613 & 0.388 & 0.410 & 0.282 & 0.392 & 0.253 \\
    & 192 & 0.426 & 0.311 & \textbf{0.377} & \textbf{0.241} & \underline{0.378} & \underline{0.253} & 0.595 & 0.342 & 0.417 & 0.276 & 0.379 & 0.256 & 0.604 & 0.373 & 0.696 & 0.379 & 0.616 & 0.382 & 0.423 & 0.287 & 0.402 & 0.258 \\
    & 336 & 0.449 & 0.320 & 0.394 & \textbf{0.250} & \textbf{0.392} & \underline{0.261} & 0.604 & 0.345 & 0.433 & 0.283 & \textbf{0.392} & 0.264 & 0.621 & 0.383 & 0.777 & 0.420 & 0.622 & 0.337 & 0.436 & 0.296 & 0.428 & 0.263 \\
    & 720 & 0.469 & 0.336 & 0.439 & \underline{0.283} & \underline{0.434} & 0.287 & 0.630 & 0.357 & 0.467 & 0.302 & \textbf{0.432} & 0.286 & 0.626 & 0.382 & 0.864 & 0.472 & 0.660 & 0.408 & 0.466 & 0.315 & 0.441 & \textbf{0.282} \\ \midrule
    \multirow{4}{*}{\rotatebox{90}{ILI}}
    & 24 & \textbf{0.930} & \textbf{0.583} & 1.351 & 0.707 & 1.514 & \underline{0.694} & 3.964 & 1.309 & 1.834 & 0.883 & \underline{1.319} & 0.754 & 3.228 & 1.260 & 4.657 & 1.449 & 2.906 & 1.182 & 2.215 & 1.081 & 1.811 & 0.823 \\
    & 36 & \textbf{1.181} & \textbf{0.669} & \underline{1.408} & \underline{0.712} & 1.519 & 0.722 & 4.095 & 1.330 & 1.742 & 0.870 & 1.579 & 0.870 & 2.679 & 1.080 & 4.650 & 1.463 & 2.585 & 1.038 & 1.963 & 0.963 & 1.763 & 0.835 \\
    & 48 & \textbf{1.283} & \textbf{0.738} & \underline{1.434} & \underline{0.721} & 1.500 & 0.725 & 4.097 & 1.326 & 1.996 & 0.815 & 1.553 & 0.815 & 2.622 & 1.078 & 5.004 & 1.542 & 3.024 & 1.145 & 2.130 & 1.024 & 1.705 & 0.818 \\
    & 60 & \textbf{1.385} & \underline{0.769} & 1.512 & 0.737 & \underline{1.418} & \textbf{0.715} & 4.355 & 1.384 & 1.806 & 0.788 & 1.470 & 0.788 & 2.857 & 1.157 & 5.071 & 1.543 & 2.761 & 1.114 & 2.368 & 1.096 & 1.708 & 0.839\\ \midrule
    \multicolumn{2}{c|}{Wins} & \textbf{33} & \textbf{27} & 1 & \underline{6} & 1 & 1 & 0 & 0 & 0 & 0 & \underline{2} & 0 & 0 & 0 & 0 & 0 & 0 & 0 & 0 & 0 & 0 & 3 \\
    \bottomrule
    \end{tabular}

\end{table}

\begin{table*}[!htbp]
    \caption{Short-term time series forecasting results. The best results are in \textbf{ bold} and the second best are in \underline{ underline}. The prediction length is set to 12.}  
    \label{tab:2}
    \centering
    \setlength{\tabcolsep}{2pt} 
    \scriptsize
    \setlength{\abovecaptionskip}{0.cm}
    \begin{tabular}{c|c|c|c|c|c|c|c|c|c|c|c|c}
    \toprule
    \multicolumn{2}{c|}{Models} & StoxLSTM & xLSTM-Mixer & P-sLSTM & iTransformer & PatchTST & Crossformer & Autoformer & FEDformer & Dlinear & TimeMixer & TimeMixer++ \\ \midrule
    \multirow{3}{*}{\rotatebox{60}{PEMS03}} 
    & MAE & \textbf{9.74} & 15.71 & 17.15 & 16.72 & 18.95 & 15.64 & 18.08 & 19.00 & 19.70 &14.63 & \underline{13.99} \\
    & MAPE & \textbf{9.59} & 14.92 & 16.56 & 15.81 & 17.29 & 15.74 & 18.75 & 18.75 & 18.35 &11.54 & \underline{13.43} \\
    & RMSE & \textbf{14.65} & 24.82 & 27.68 & 27.81 & 30.15 & 25.56 & 27.82 & 30.05 & 32.35 &23.28 & \underline{24.03} \\ \midrule
    \multirow{3}{*}{\rotatebox{60}{PEMS04}} 
    & MAE & \textbf{14.59} & 24.61 & 23.01 & 21.81 & 24.86 & 20.38 & 25.00 & 26.51 & 24.62 &19.21 & \underline{17.46} \\
    & MAPE & \textbf{9.25} & 15.48 & 14.44 & 13.42 & 16.65 & 12.84 & 16.70 & 16.76 & 16.12 &12.53 & \underline{11.34} \\
    & RMSE & \textbf{23.48} & 38.92 & 35.92 & 33.91 & 40.46 & 32.41 & 38.02 & 41.82 & 39.51 &30.92 & \underline{28.73} \\ \midrule
    \multirow{3}{*}{\rotatebox{60}{PEMS07}} 
    & MAE & \textbf{14.36} & 27.98 & 23.54 & 23.01 & 27.87 & 22.79 & 26.92 & 27.92 & 28.65 & 20.57 & \underline{18.38} \\
    & MAPE & \textbf{6.32} & 12.20 & 9.78 & 10.02 & 12.69 & 9.41 & 11.83 & 12.29 & 12.15 & 8.62 & \underline{7.32} \\
    & RMSE & \textbf{22.34} & 43.92 & 37.70 & 35.56 & 42.56 & 35.61 &40.60 & 42.29 & 45.02 & 33.59 & \underline{31.75} \\ \midrule
    \multirow{3}{*}{\rotatebox{60}{PEMS08}} 
    & MAE & \textbf{12.55} & 16.56 & 17.20 & 17.94 & 20.35 & 17.38 & 20.47 & 20.56 & 20.26 &15.22 & \underline{13.81} \\
    & MAPE & \textbf{7.79} & 10.24 & 10.87 & 10.93 & 13.15 & 10.80 & 12.27 & 12.41 & 12.09 &9.67 & \underline{8.21} \\
    & RMSE & \textbf{19.45} & 26.56 & 27.32 & 27.88 & 31.04 & 27.34 & 31.52 & 32.97 & 32.38 &24.26 & \underline{23.62} \\
    \bottomrule
    \end{tabular}

\end{table*}

In long-term forecasting tasks, xLSTM-based models generally demonstrate superior performance compared to Transformer-based and Linear-based models, highlighting the strength of recurrent architectures like xLSTM in capturing long-range temporal dependencies. Among these, StoxLSTM achieves the best results. For instance, on the Weather dataset, StoxLSTM attains an average MSE of 0.151 across different prediction horizons, which is \textbf{31.1\%} lower than the xLSTM-Mixer’s average MSE of 0.219. Similarly, on the Electricity dataset, StoxLSTM achieves an average MSE of 0.139, outperforming xLSTM-Mixer’s 0.153 by \textbf{9.1\%}, further demonstrating its consistent advantage. Notably, StoxLSTM maintains this leading performance across other datasets such as Solar, ETTh1, and ETTm1, reflecting its robustness and generalizability in diverse long-term forecasting scenarios.

In short-term forecasting, although Linear-based models show competitive results, StoxLSTM still surpasses them by a significant margin. Specifically, across four traffic datasets (PEMS03, PEMS04, PEMS07, and PEMS08), StoxLSTM achieves an average MAE of 12.81, outperforming the best Linear-based model, TimeMixer++, which records an MAE of 15.91, representing a \textbf{19.49\%} improvement. Furthermore, StoxLSTM attains an average MAPE of 8.24, which is \textbf{18.25\%} lower than TimeMixer++’s 10.08. In terms of RMSE, StoxLSTM achieves 19.98 on average, outperforming TimeMixer++’s 27.03 by \textbf{26.08\%}. These consistent improvements across multiple metrics and datasets underscore StoxLSTM’s effectiveness in capturing short-term temporal patterns with high accuracy.  

\begin{figure}[!htbp]
    \centering
    \begin{subfigure}{0.25\textwidth}
        \centering
        \includegraphics[width=\textwidth]{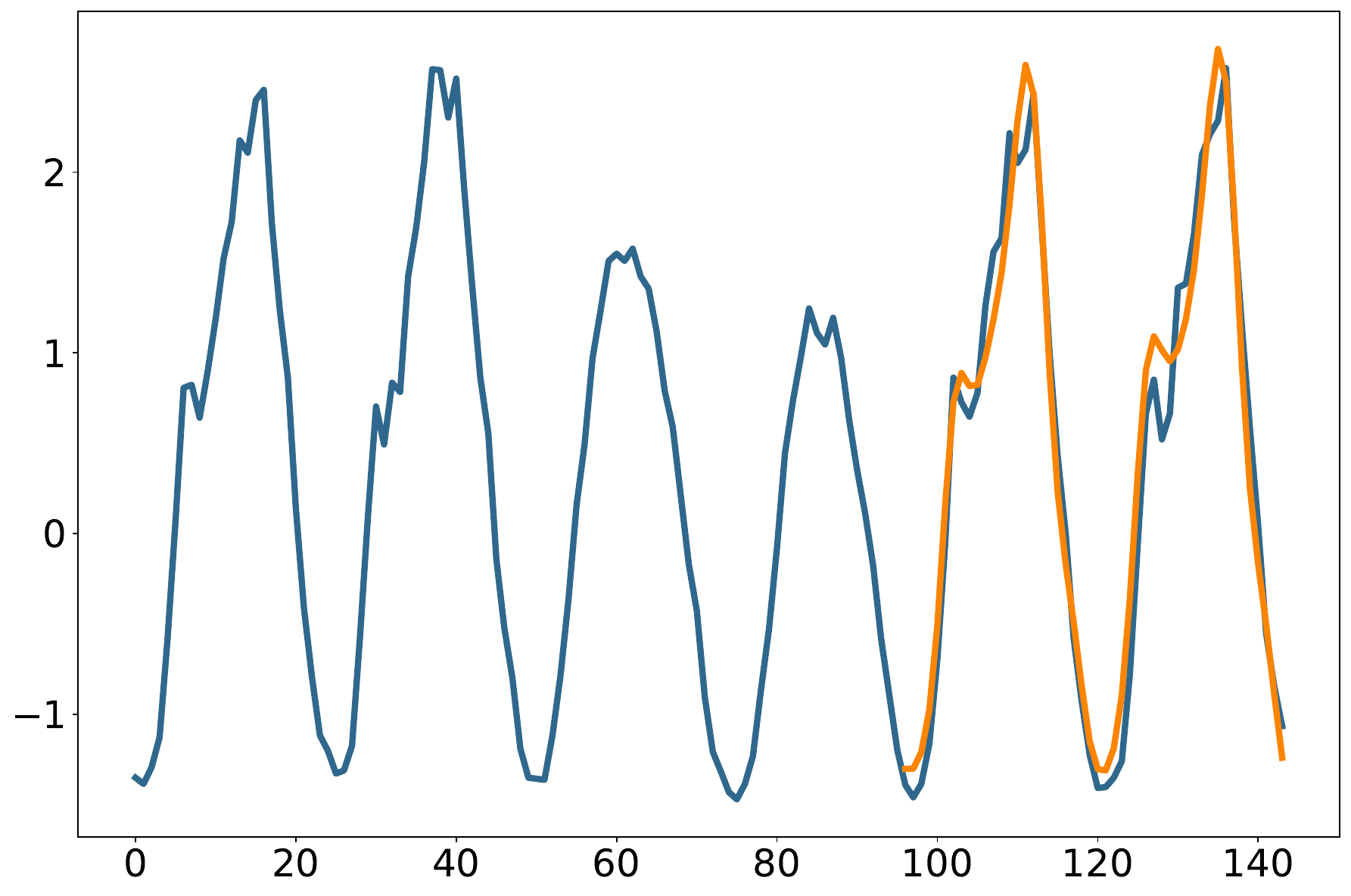} 
        \caption{Horizon: 48}
        \label{traffic_48}
    \end{subfigure}\hspace{-0.1cm}
    \begin{subfigure}{0.25\textwidth}
        \centering
        \includegraphics[width=\textwidth]{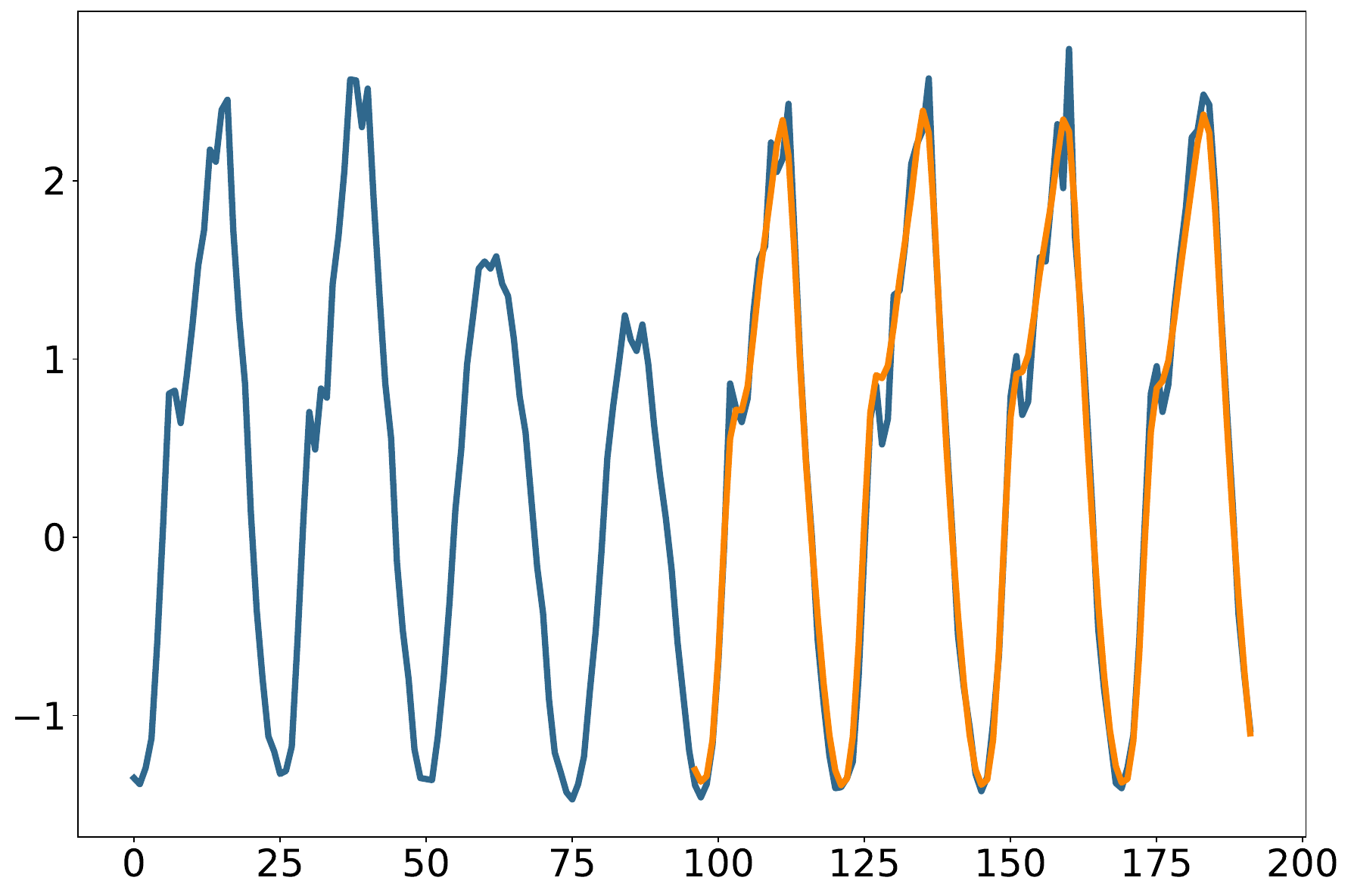} 
        \caption{Horizon: 96}
        \label{traffic_96}
    \end{subfigure}\hspace{-0.1cm}
    \begin{subfigure}{0.25\textwidth}
        \centering
        \includegraphics[width=\textwidth]{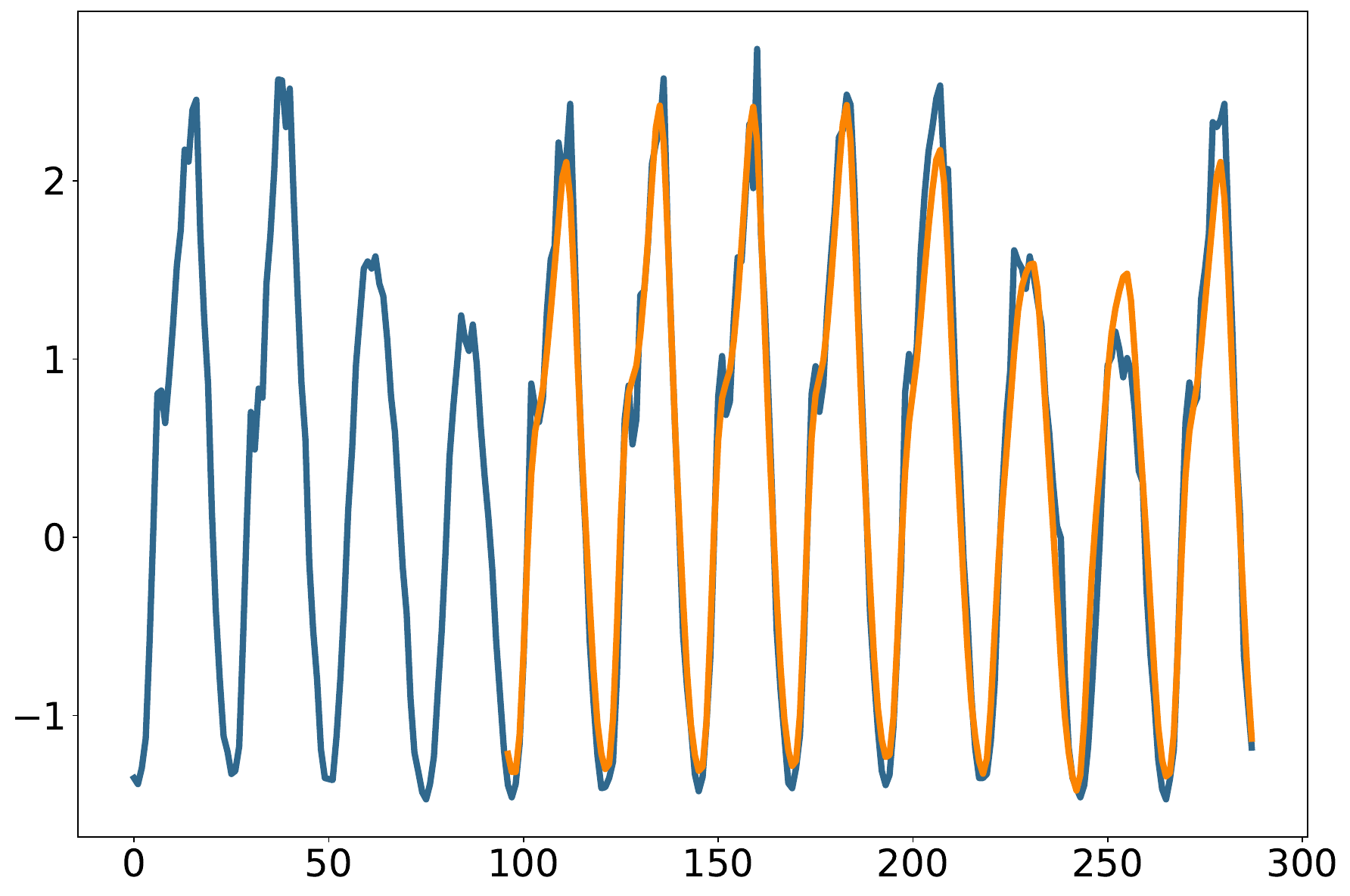} 
        \caption{Horizon: 192}
        \label{traffic_192}
    \end{subfigure}\hspace{-0.1cm}
    \begin{subfigure}{0.25\textwidth}
        \centering
        \includegraphics[width=\textwidth]{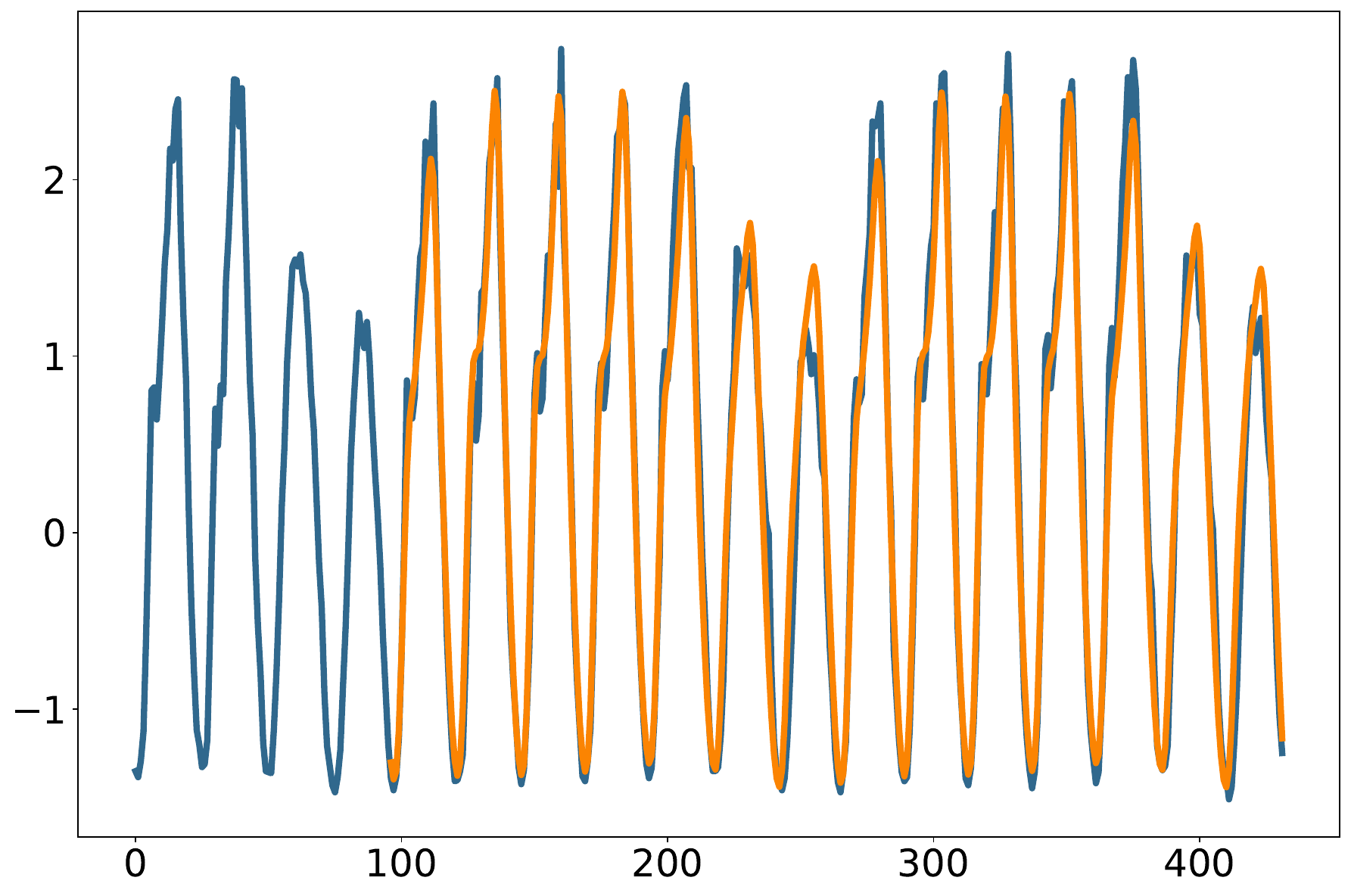} 
        \caption{Horizon: 336}
        \label{traffic_336}
    \end{subfigure}\hspace{-0.1cm}
    \caption{Visualization of the prediction results from StoxLSTM with different prediction horizons H=\{48, 96, 192, 336\} from the Traffic dataset. The orange line denotes the prediction values generated by StoxLSTM, while the blue line represents the ground truth.}
    \label{fig:Traffic}
\end{figure}

\begin{figure}[!htbp]
    \centering
    \begin{subfigure}{0.25\textwidth}
        \centering
        \includegraphics[width=\textwidth]{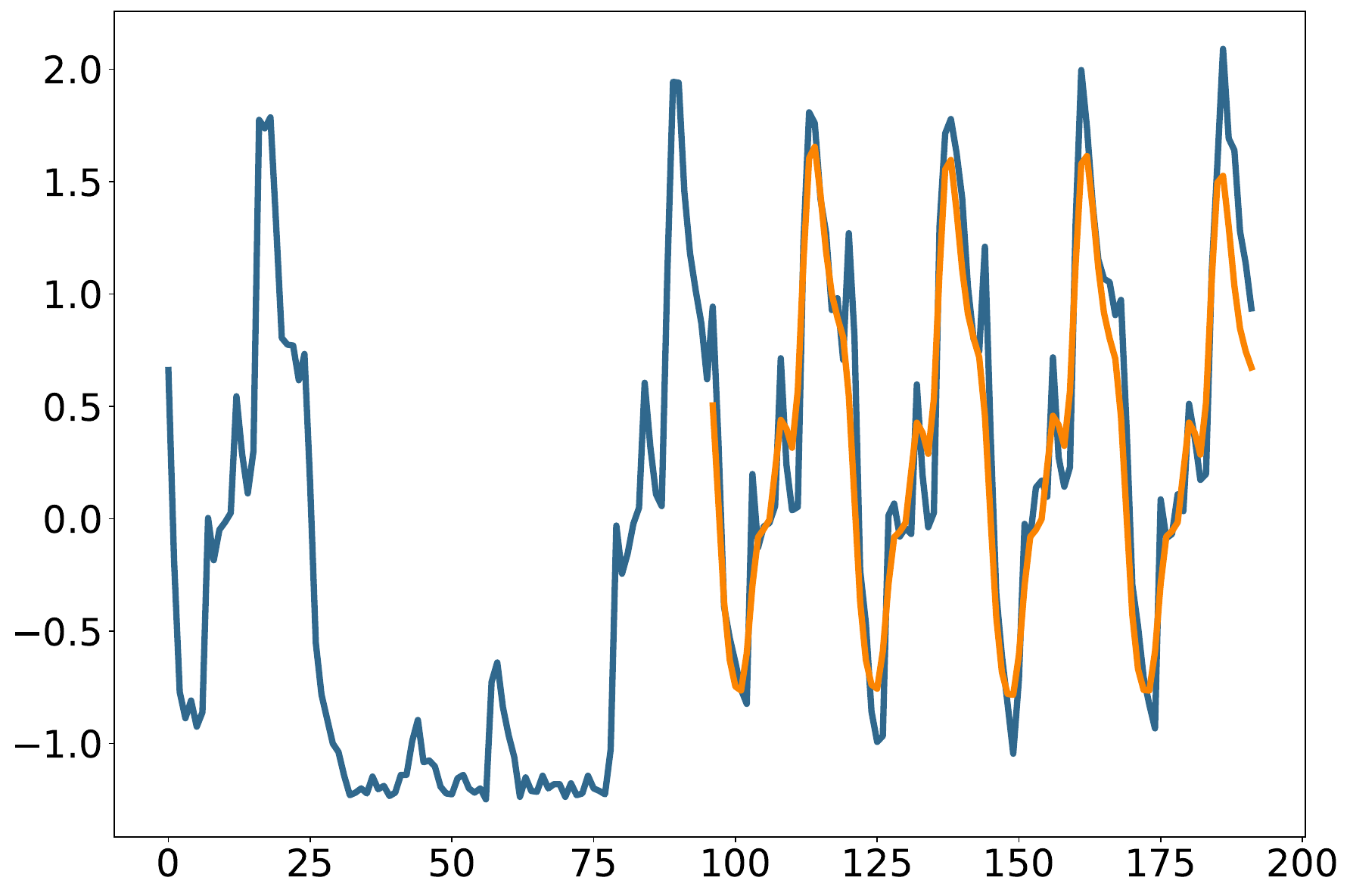} 
        \caption{Horizon: 96}
        \label{ECL_96}
    \end{subfigure}\hspace{-0.1cm}
    \begin{subfigure}{0.25\textwidth}
        \centering
        \includegraphics[width=\textwidth]{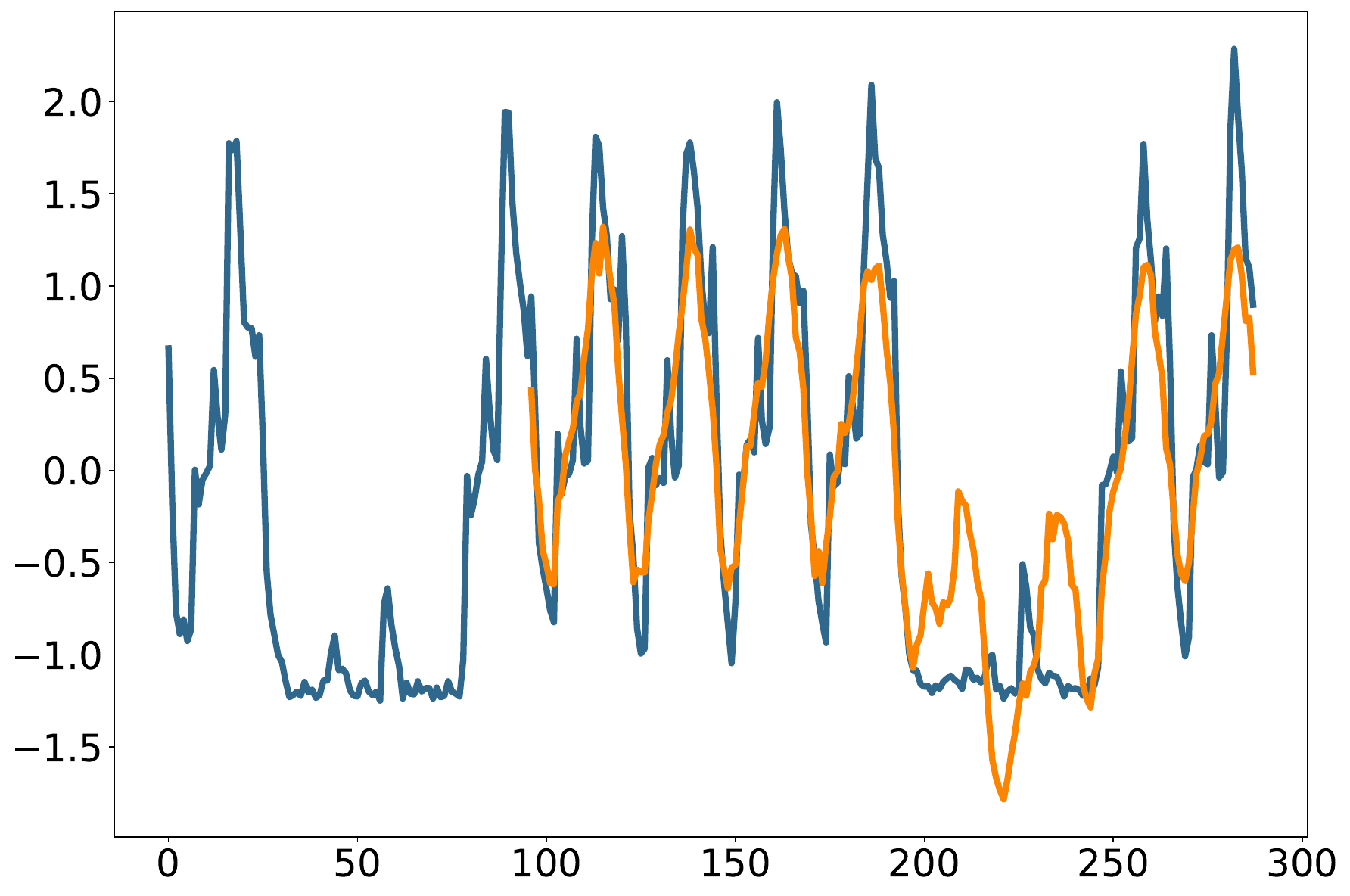} 
        \caption{Horizon: 192}
        \label{ECL_192}
    \end{subfigure}\hspace{-0.1cm}
    \begin{subfigure}{0.25\textwidth}
        \centering
        \includegraphics[width=\textwidth]{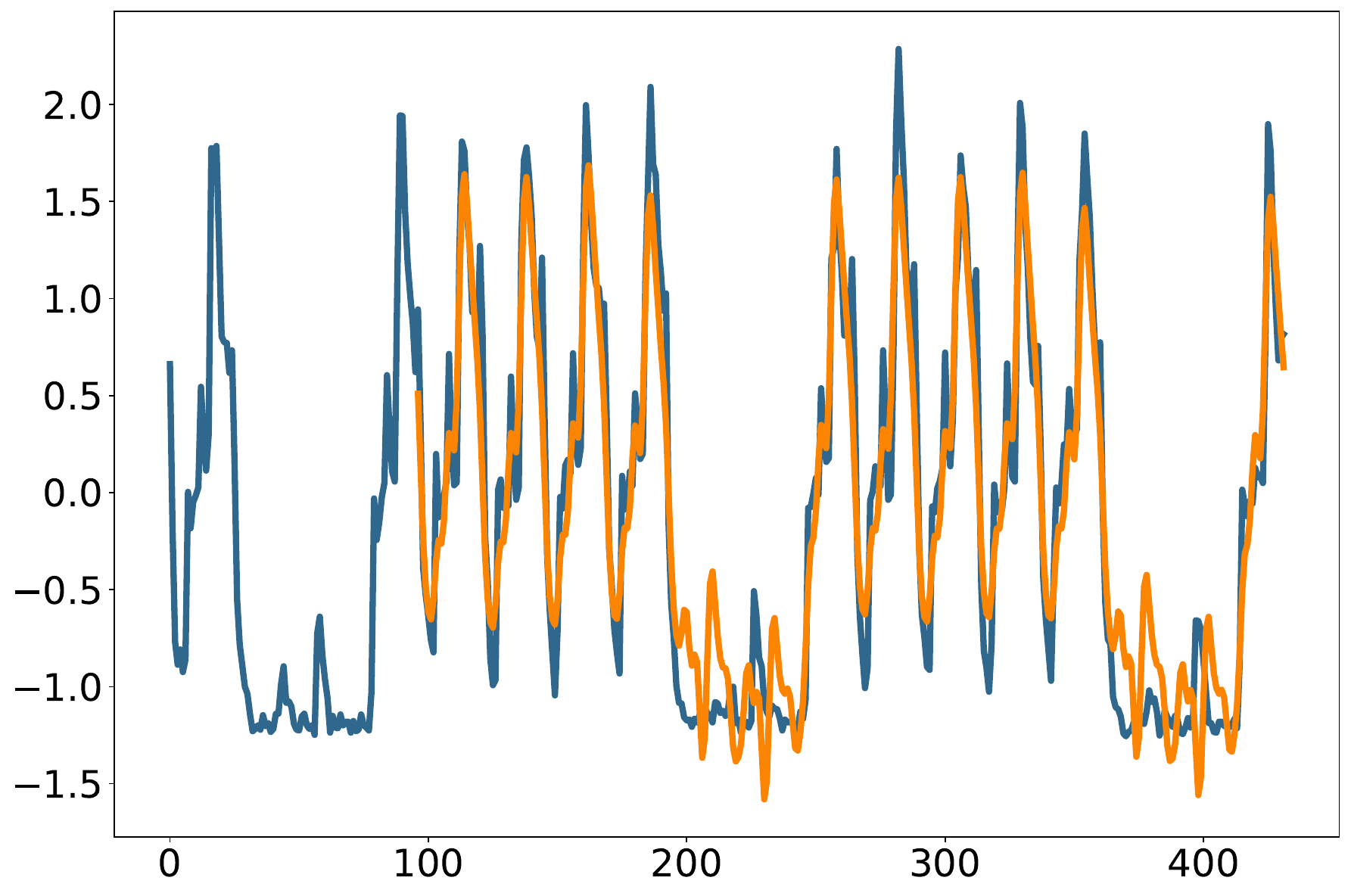} 
        \caption{Horizon: 336}
        \label{ECL_336}
    \end{subfigure}\hspace{-0.1cm}
    \begin{subfigure}{0.25\textwidth}
        \centering
        \includegraphics[width=\textwidth]{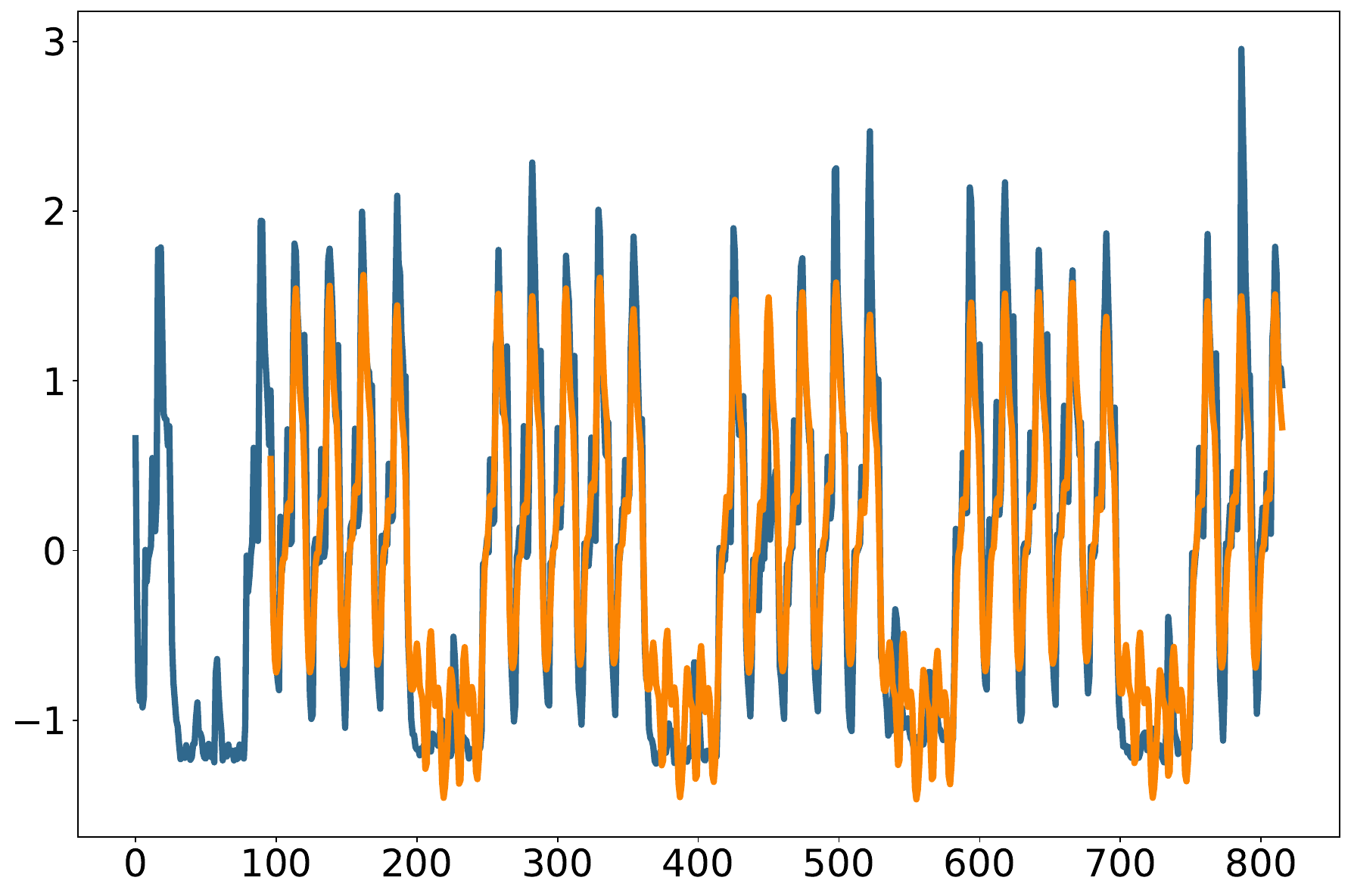} 
        \caption{Horizon: 720}
        \label{ECL_1720}
    \end{subfigure}\hspace{-0.1cm}
    \caption{Visualization of the prediction results from StoxLSTM with different prediction horizons H=\{96, 192, 336, 720\} from the Electricity dataset. The orange line denotes the prediction values generated by StoxLSTM, while the blue line represents the ground truth.}
    \label{fig:ECL}
\end{figure}

\begin{figure}[!htbp]
    \centering
    \begin{subfigure}{0.25\textwidth}
        \centering
        \includegraphics[width=\textwidth]{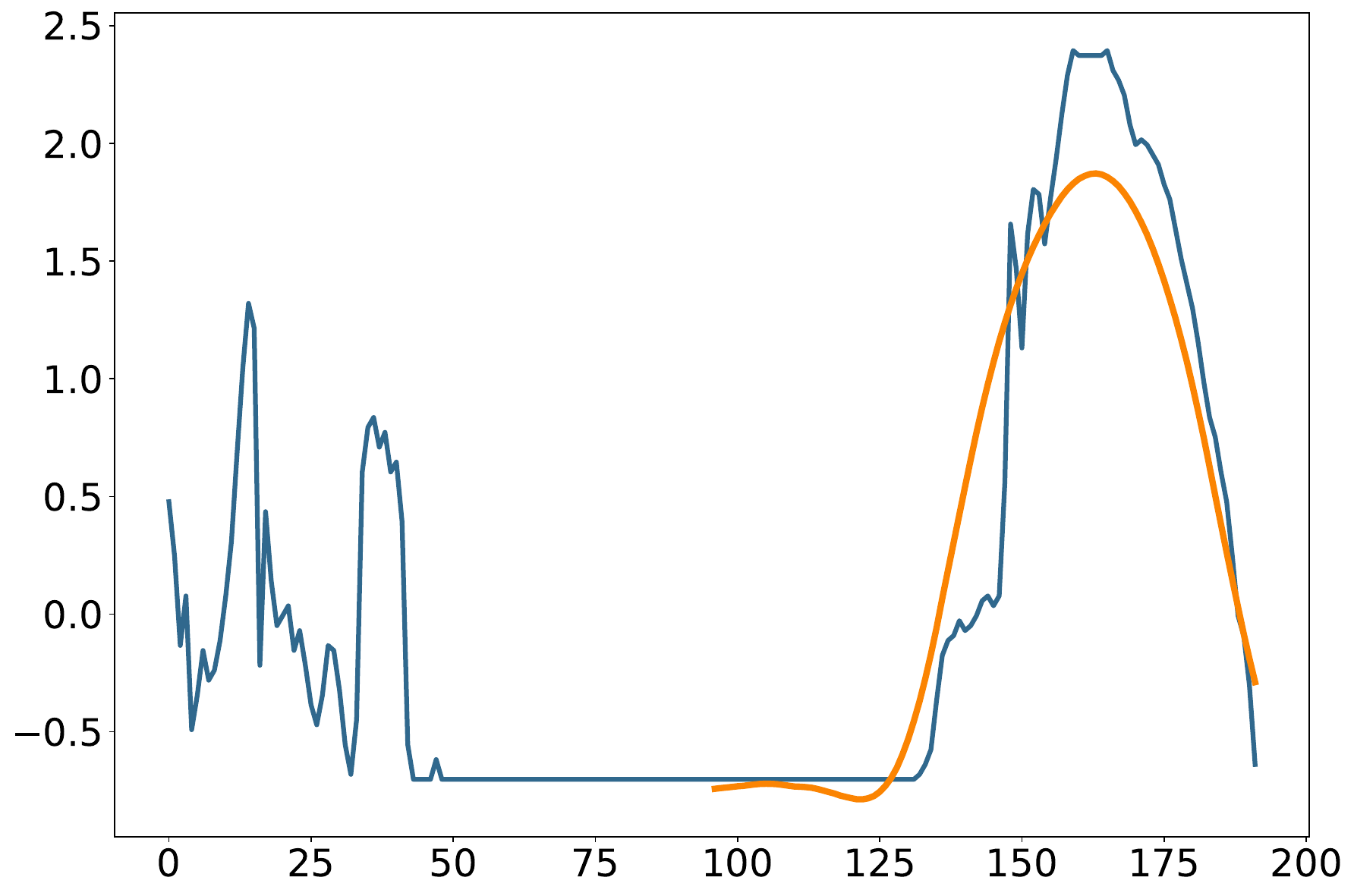} 
        \caption{Horizon: 96}
        \label{solar_96}
    \end{subfigure}\hspace{-0.1cm}
    \begin{subfigure}{0.25\textwidth}
        \centering
        \includegraphics[width=\textwidth]{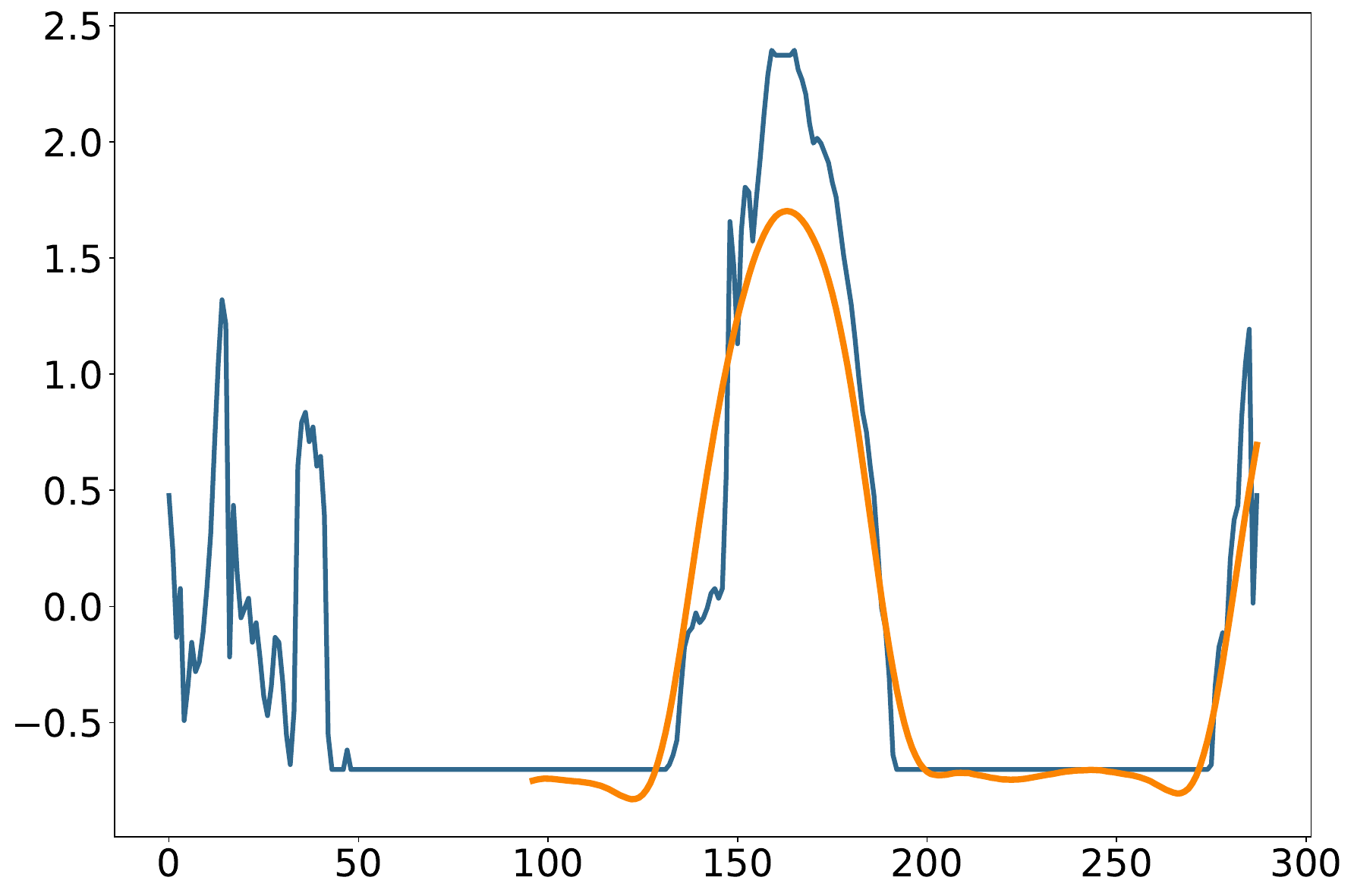} 
        \caption{Horizon: 192}
        \label{solar_192}
    \end{subfigure}\hspace{-0.1cm}
    \begin{subfigure}{0.25\textwidth}
        \centering
        \includegraphics[width=\textwidth]{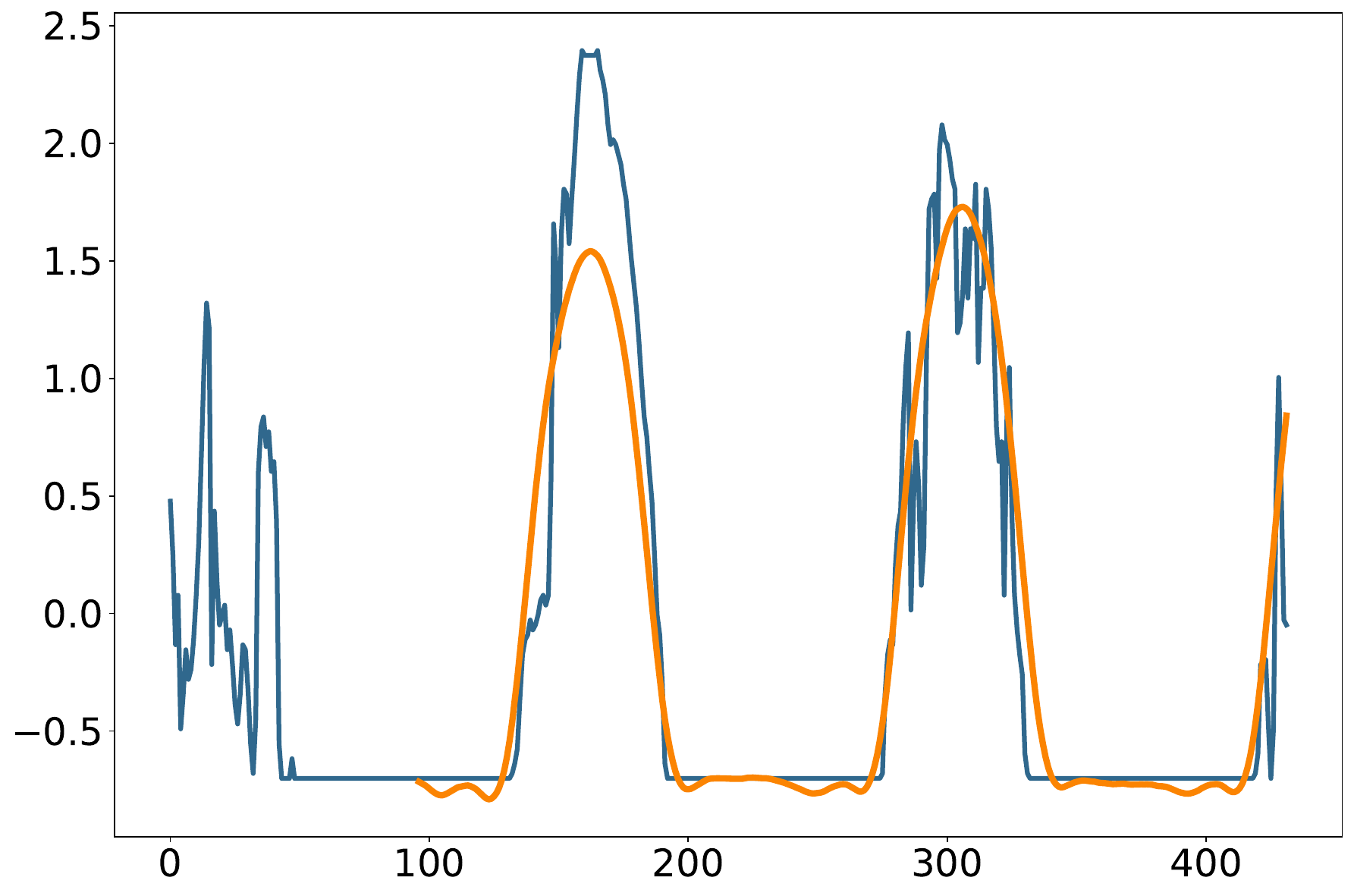} 
        \caption{Horizon: 336}
        \label{solar_336}
    \end{subfigure}\hspace{-0.1cm}
    \begin{subfigure}{0.25\textwidth}
        \centering
        \includegraphics[width=\textwidth]{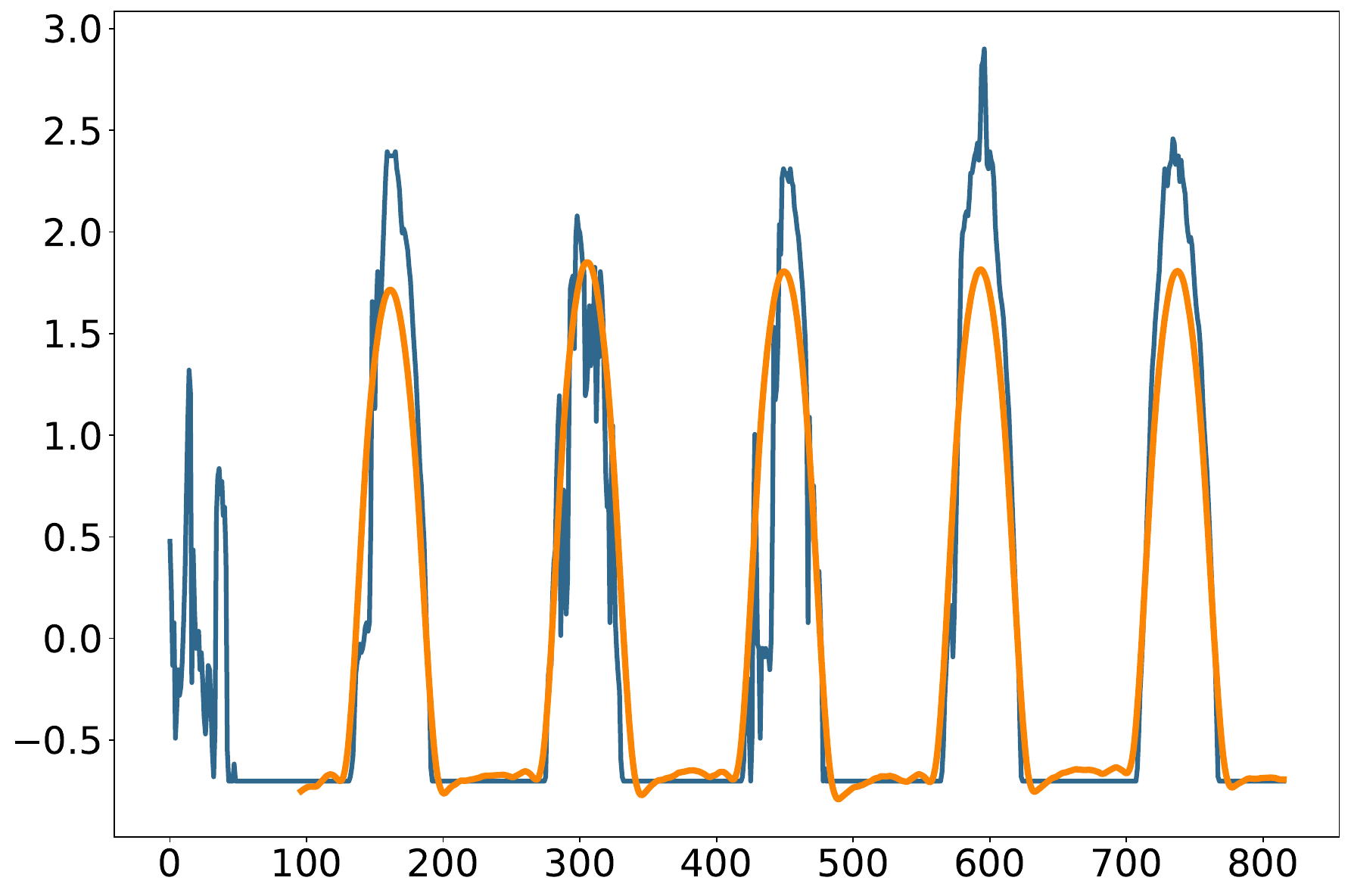} 
        \caption{Horizon: 720}
        \label{solar_720}
    \end{subfigure}\hspace{-0.1cm}
    \caption{Visualization of the prediction results from StoxLSTM with different prediction horizons H=\{96, 192, 336, 720\} from the Solar dataset. The orange line denotes the prediction values generated by StoxLSTM, while the blue line represents the ground truth.}
    \label{fig:Solar}
\end{figure}

\begin{figure}[!htbp]
    \centering
    \begin{subfigure}{0.25\textwidth}
        \centering
        \includegraphics[width=\textwidth]{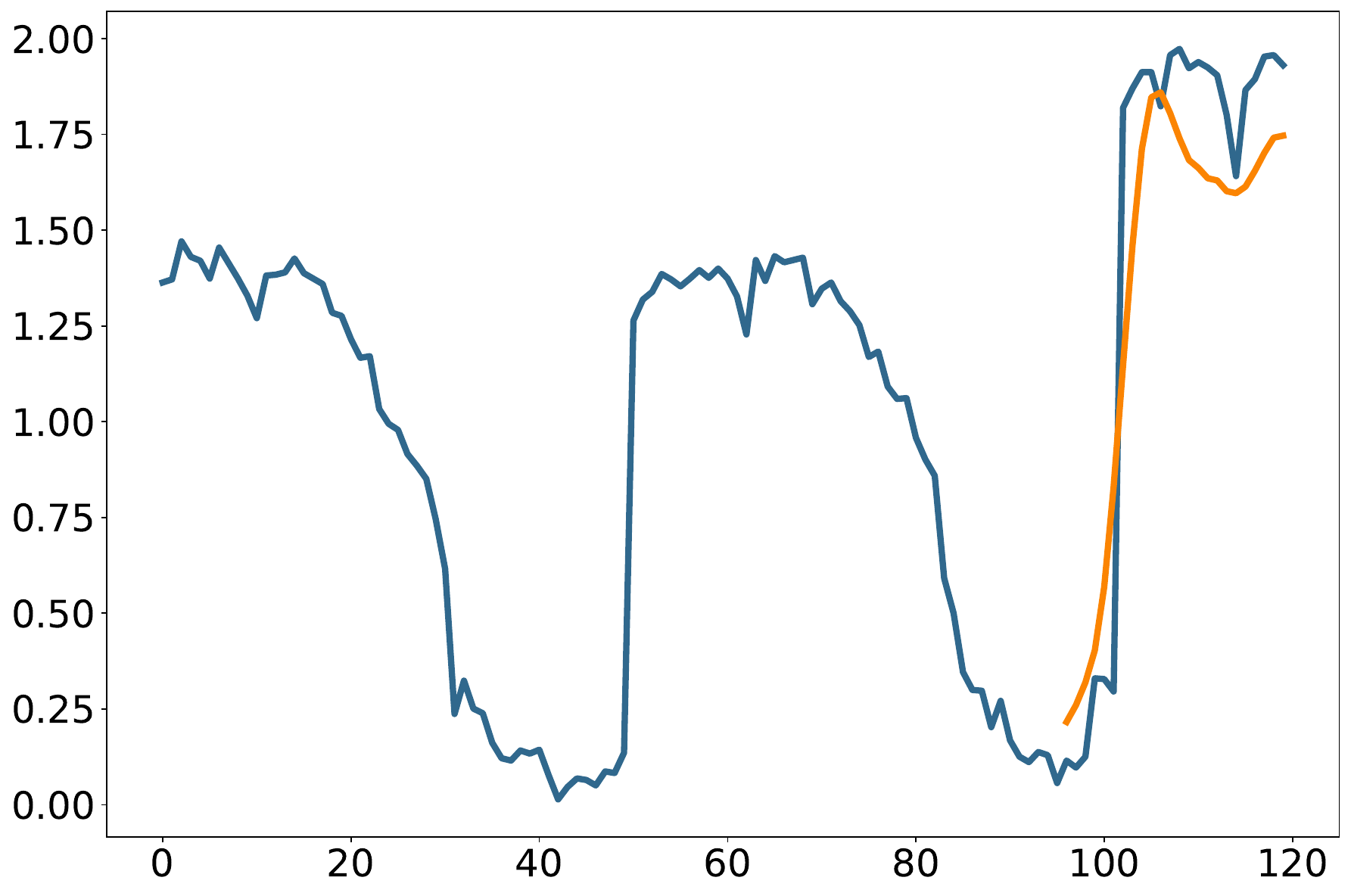} 
        \caption{Horizon: 24}
        \label{ILI_24}
    \end{subfigure}\hspace{-0.1cm}
    \begin{subfigure}{0.25\textwidth}
        \centering
        \includegraphics[width=\textwidth]{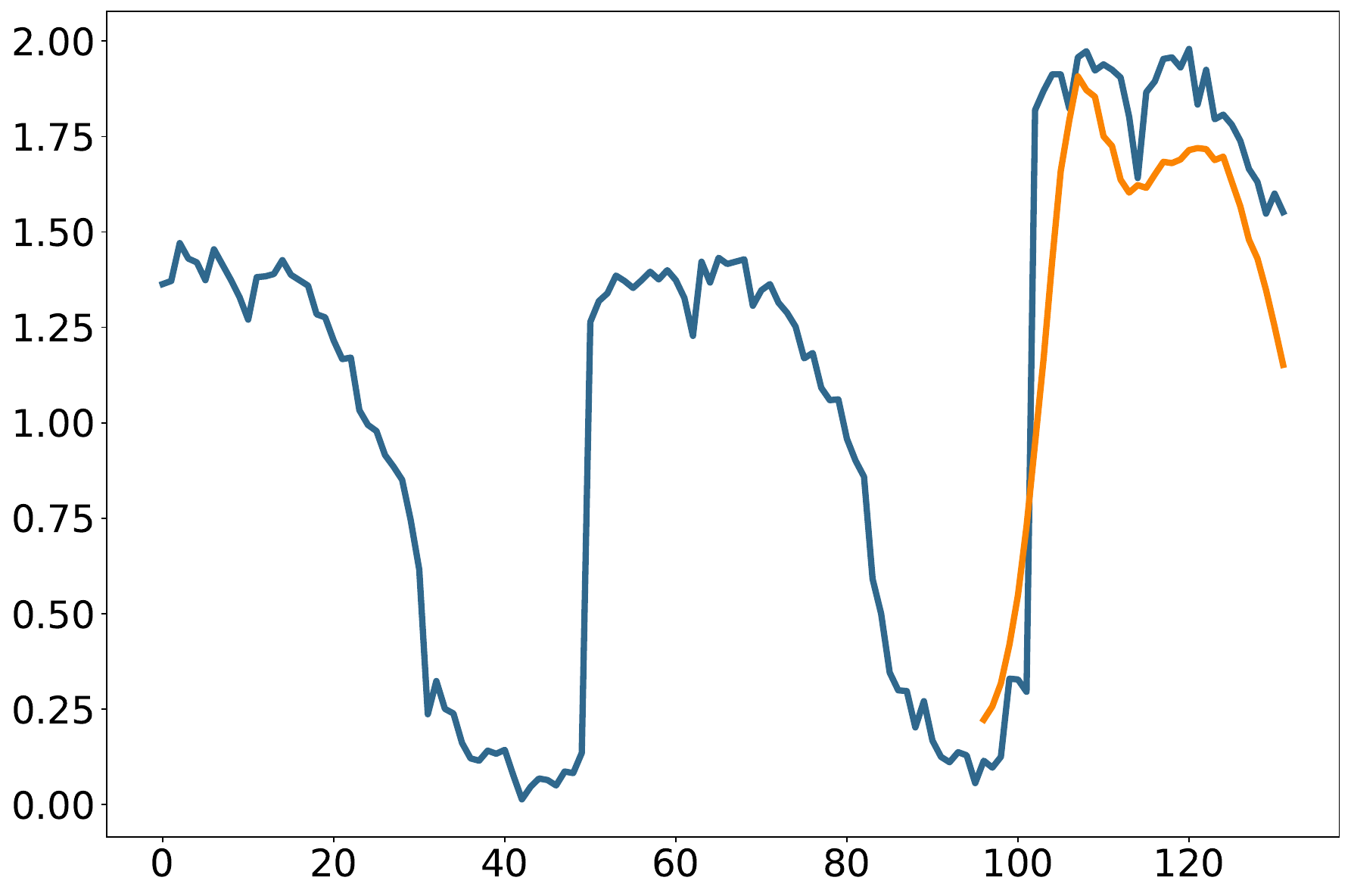} 
        \caption{Horizon: 36}
        \label{ILI_36}
    \end{subfigure}\hspace{-0.1cm}
    \begin{subfigure}{0.25\textwidth}
        \centering
        \includegraphics[width=\textwidth]{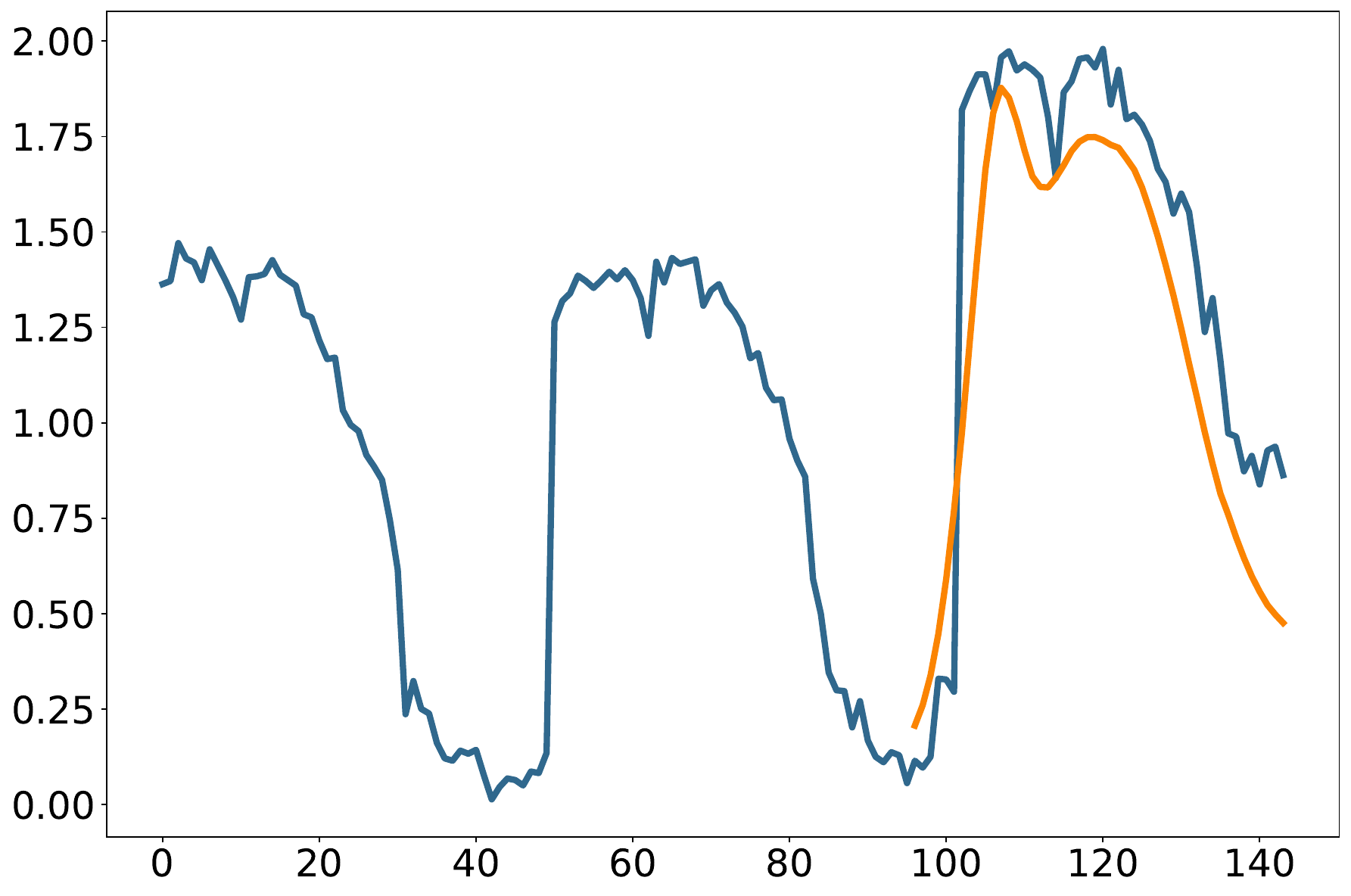} 
        \caption{Horizon: 48}
        \label{ILI_48}
    \end{subfigure}\hspace{-0.1cm}
    \begin{subfigure}{0.25\textwidth}
        \centering
        \includegraphics[width=\textwidth]{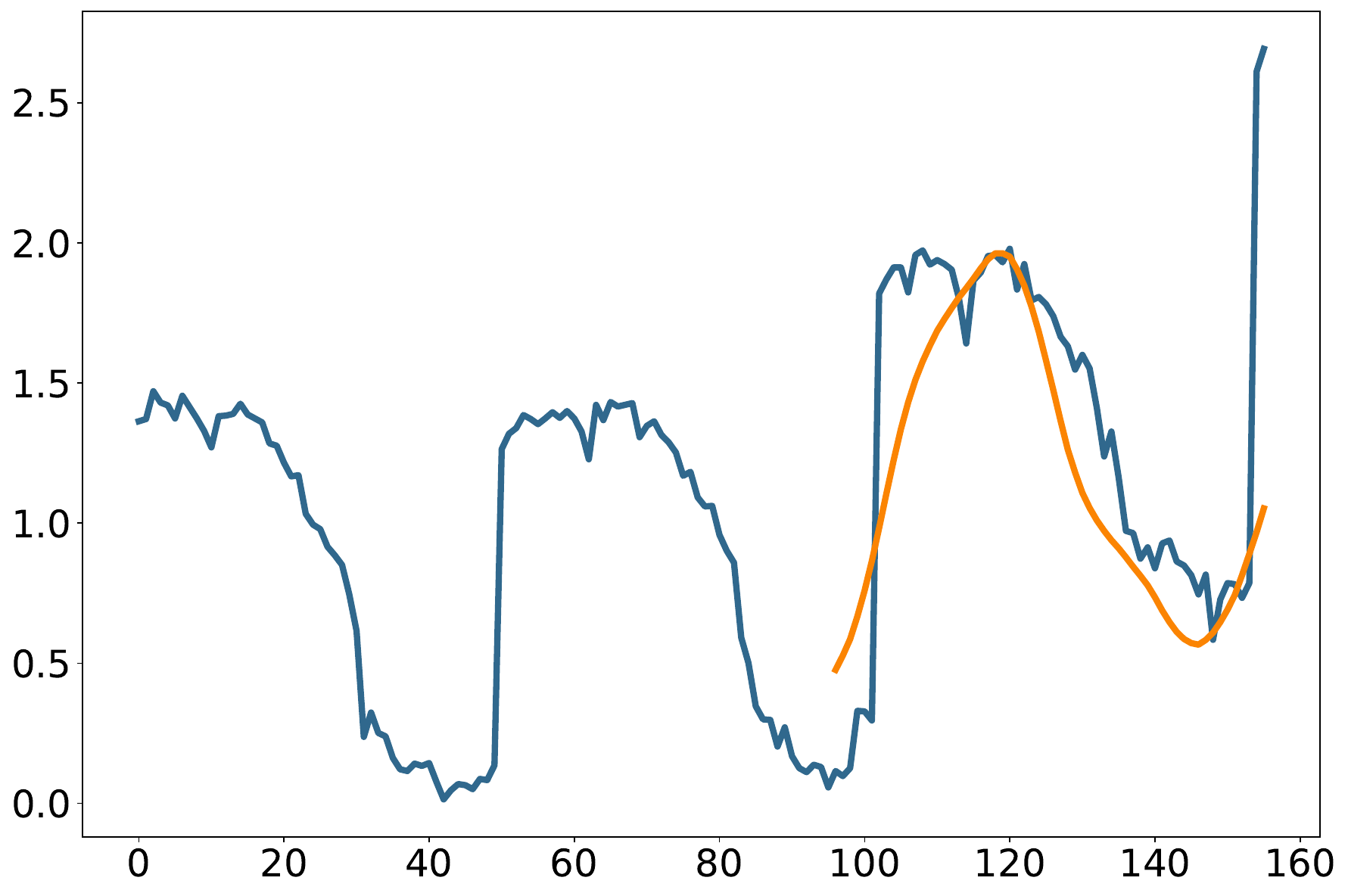} 
        \caption{Horizon: 60}
        \label{ILI_60}
    \end{subfigure}\hspace{-0.1cm}
    \caption{Visualization of the prediction results from StoxLSTM with different prediction horizons H=\{24, 36, 48, 60\} from the ILI dataset. The orange line denotes the prediction values generated by StoxLSTM, while the blue line represents the ground truth.}
    \label{fig:ILI}
\end{figure}

Figs. \ref{fig:Traffic} to \ref{fig:ILI} present partial visualizations of StoxLSTM’s prediction results. It can be observed that the prediction curves generated by StoxLSTM tend to be relatively smooth, and its ability to capture fine-grained variations in time series appears somewhat limited. This phenomenon could potentially be ascribed to the fact that StoxLSTM adopts a variational inference framework analogous to that of VAEs \cite{Kingma2013AutoEncodingVB,girin_Dynamical_2021} for learning the characteristics of latent variables.



Overall, these results demonstrate that StoxLSTM consistently delivers state-of-the-art accuracy across both short- and long-term forecasting tasks, validating its effectiveness and robustness in diverse time series forecasting scenarios.

\subsection{Model Analysis}
\subsubsection{Ablation Study.}

We study the effect of patching combined with channel independence (P + CI), series decomposition (SD), stochastic latent variables (stochastic), and the SSM structure on forecasting performance, with results summarized in Table \ref{tab:4}. P-sLSTM is employed as a baseline to compare against the SSM design integrated within StoxLSTM. For the model without patching and channel independence, the look-back length is reduced to 96 steps (compared to 336 steps for the other models) to prevent overfitting and reduce training time and memory consumption. Additionally, xLSTM-Mixer is included as the state-of-the-art benchmark for xLSTM-based models.

\begin{table}[!htbp]
    \caption{Ablation studies on StoxLSTM are conducted with 5 variants: (a) original StoxLSTM; (b) StoxLSTM without stochastic latent variables, where the latent state \( z_t \) is generated deterministically without random Gaussian sampling (w/o stochastic); (c) StoxLSTM without patching and channel independence methods (w/o P + CI); (d) StoxLSTM without the series decomposition block (w/o SD); and (e) P-sLSTM, which incorporates P + CI and SD but does not apply the SSM enhancement to the xLSTM architecture.}
    \label{tab:4}
    \centering
    \scriptsize
    \setlength{\abovecaptionskip}{0.cm}
    \begin{tabular}{c|c|cc|cc|cc|cc|cc||cc}
    \toprule
    \multicolumn{2}{c|}{Models} & \multicolumn{2}{c|}{StoxLSTM} & \multicolumn{2}{c|}{w/o stochastic} & \multicolumn{2}{c|}{w/o P + CI} & \multicolumn{2}{c|}{w/o SD} & \multicolumn{2}{c||}{P-sLSTM} & \multicolumn{2}{c}{xLSTM-Mixer} \\ \midrule
    \multicolumn{2}{c|}{Metric} & MSE & MAE & MSE & MAE & MSE & MAE & MSE & MAE & MSE & MAE & MSE & MAE\\ \midrule
    \multirow{4}{*}{\rotatebox{90}{Solar}} 
    & 96 & \underline{0.098} & \underline{0.179} & 0.106 & 0.189 & \textbf{0.095} & \textbf{0.174} & 0.119 & 0.201 & 0.167 & 0.232 & 0.227 & 0.313 \\
    & 192 & \underline{0.124} & \underline{0.212} & 0.129 & 0.218 & \textbf{0.119} & \textbf{0.206} & 0.136 & 0.226 & 0.180 & 0.241 & 0.245 &	0.328 \\
    & 336 & \textbf{0.139} & \textbf{0.224} & \underline{0.144} & \underline{0.231} & 0.159 & 0.248 & 0.152 & 0.239 & 0.190 & 0.248 &	0.255 &	0.330 \\
    & 720 & \textbf{0.169} & \textbf{0.241} & \underline{0.171} & \underline{0.247} & 0.250 & 0.320 & 0.173 & 0.248 & 0.196 & 0.249 &	0.257 &	0.324  \\ \midrule
    \multirow{4}{*}{\rotatebox{90}{ETTm1}}
    & 96 & \textbf{0.223} & \textbf{0.310} & \underline{0.229} & \underline{0.315} & 0.304 & 0.375 & 0.253 & 0.336 & 0.292 & 0.343 & 0.275 & 0.184  \\
    & 192 & \textbf{0.261} & \textbf{0.339} & \underline{0.265} & \underline{0.340} & 0.494 & 0.461 & 0.275 & 0.351 & 0.329 & 0.369 & 0.319 & 0.354  \\
    & 336 & \textbf{0.290} & \textbf{0.360} & 0.293 & 0.362 & 0.592 & 0.499 & \textbf{0.290} & \textbf{0.360} & 0.362 & 0.391 & 0.353 & 0.370 \\
    & 720 & \textbf{0.328} & \textbf{0.385} & 0.332 & 0.391 & 0.645 & 0.526 & \textbf{0.328} & \textbf{0.385} & 0.421 & 0.424 & 0.409 & 0.407 \\ \midrule
    \multirow{4}{*}{\rotatebox{90}{Electricity}}
    & 96 & \textbf{0.117} & \underline{0.223} & \underline{0.120} & 0.229 & 0.357 & 0.421 &0.151 & 0.260 & 0.130 & 0.226 & 0.126 & \textbf{0.218} \\
    & 192 & \textbf{0.136} & \underline{0.242} & \textbf{0.136} & 0.244 & 0.454 & 0.498 & 0.152 & 0.261 & 0.148 & 0.243 & 0.144 & \textbf{0.235} \\
    & 336 & \textbf{0.144} & \underline{0.253} & \underline{0.146} & 0.255 & 0.621 & 0.608 & 0.158 & 0.268 & 0.165 & 0.262 & 0.157 & \textbf{0.250} \\
    & 720 & \textbf{0.159} & \textbf{0.270} & \underline{0.161} & \underline{0.274} & 0.699 & 0.668 & 0.162 & 0.278 & 0.199 & 0.293 & 0.183 & 0.276 \\ \bottomrule
    \end{tabular}
\end{table}

Experimental results clearly demonstrate that all components—series decomposition, patching, channel independence, stochastic latent variables, and the SSM structure—contribute positively to the overall performance of StoxLSTM. Focusing first on the SSM modeling aspect, both the original StoxLSTM and its variants without stochastic latent variables (w/o Stochastic) and without series decomposition (w/o SD) consistently outperform P-sLSTM across all experiments, achieving lower MSE and MAE. This confirms that the state space model enhancement to the xLSTM architecture plays a crucial role in improving forecasting performance. Moreover, StoxLSTM with stochastic latent variables generally achieves better and more stable prediction results. Notably, the Solar-Energy and ETTm1 datasets exhibit higher randomness, with forecastability values \cite{wang_timemixerpuls_2024, goerg_forecastable_2013} of 0.33 and 0.46, respectively, while the Electricity dataset shows lower randomness, with a forecastability of 0.77. The performance improvements attributed to stochastic latent variables are more pronounced on the Solar-Energy and ETTm1 datasets compared to the Electricity dataset. This demonstrates that modeling latent states as stochastic variables effectively captures the inherent randomness in time series, enabling better forecasting accuracy and robustness, particularly for datasets with stronger stochasticity. This indicates that modeling latent states as stochastic variables effectively captures the inherent randomness in time series, leading to enhanced forecasting accuracy and robustness, especially for datasets with stronger stochasticity. In contrast, the model variant without patching and channel independence exhibits notable instability: while it achieves excellent results in certain scenarios, its performance deteriorates significantly in others. This suggests that patching and channel independence not only improve robustness but also enable the model to effectively handle longer look-back lengths under constrained computational and memory resources, thereby reducing the risk of overfitting.

\subsubsection{Hyperparameter sensitivity}
\begin{figure}[!htbp]
  \centering 
\includegraphics[width=1\textwidth]{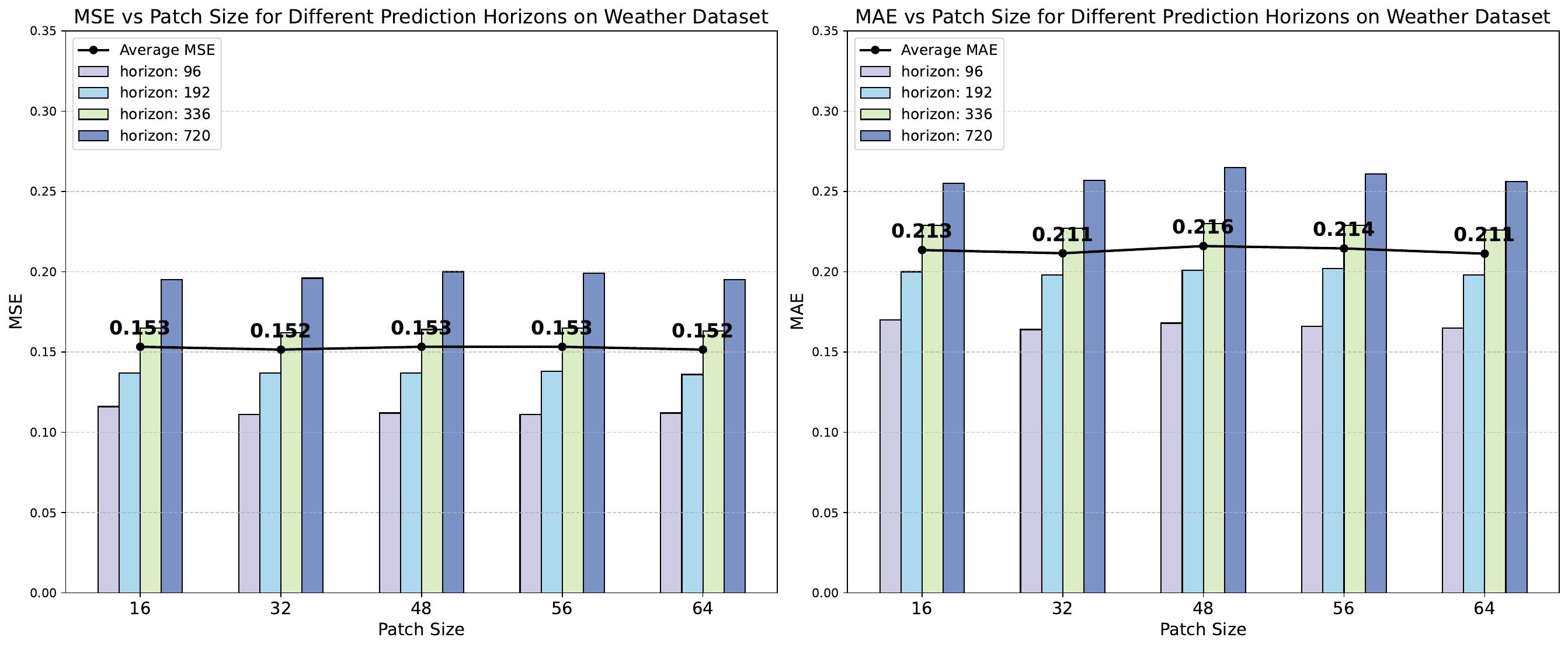} 
  \caption{Impact of different patch sizes on prediction performance on the Weather dataset. Colors correspond to different prediction horizons, and the black line represents the average MSE or MAE.}
  \label{error_vs_patchsize_Weather}
\end{figure}

\begin{figure}[!htbp]
  \centering 
\includegraphics[width=1\textwidth]{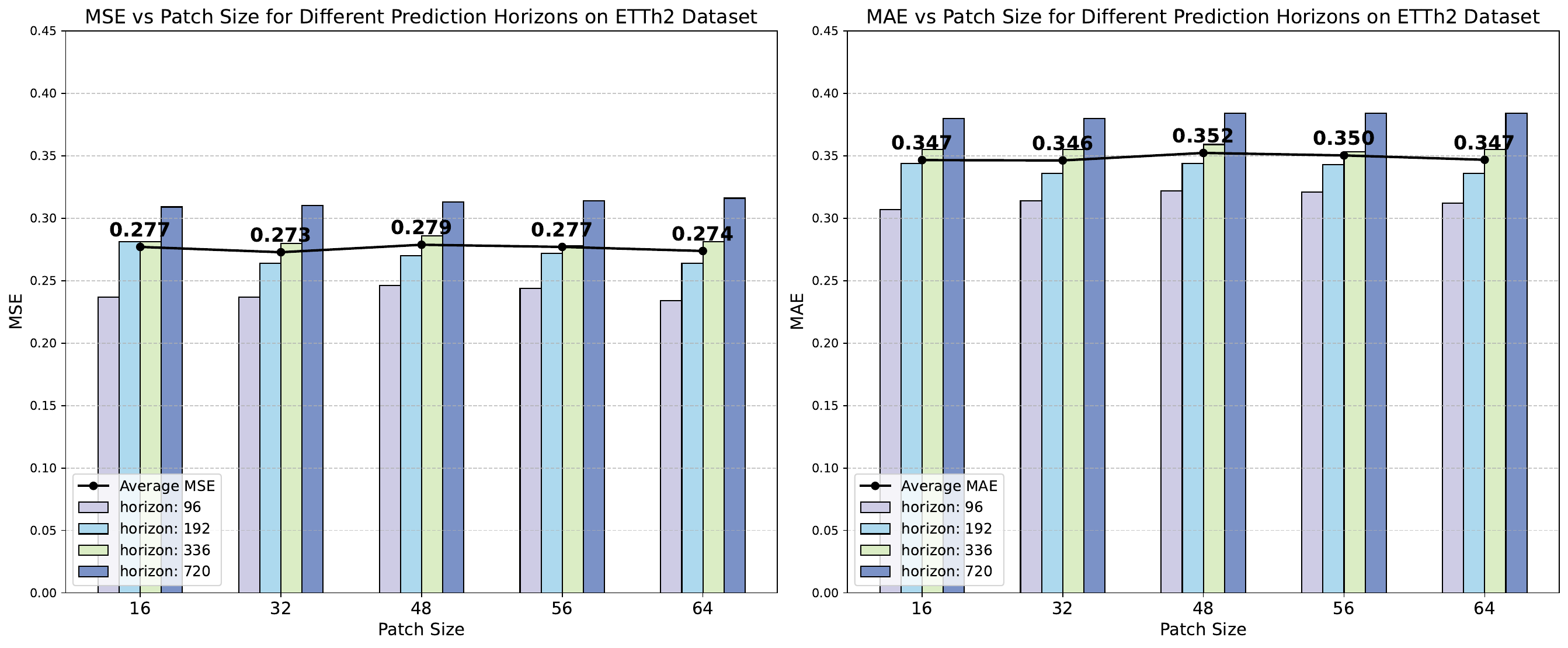} 
  \caption{Impact of different patch sizes on prediction performance on the ETTh2 dataset. Colors correspond to different prediction horizons, and the black line represents the average MSE or MAE.}
  \label{error_vs_patchsize_ETTh2}
\end{figure}

\begin{figure}[!htbp]
  \centering 
\includegraphics[width=1\textwidth]{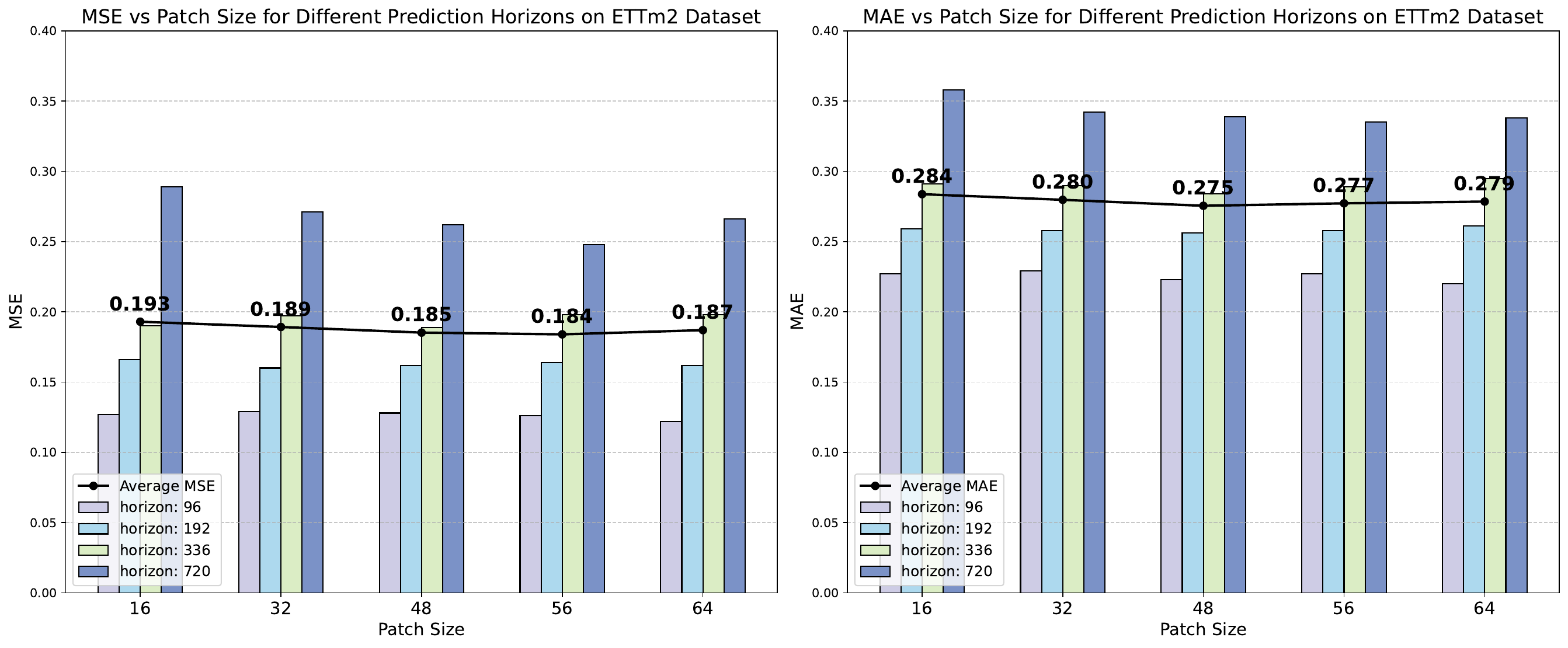} 
  \caption{Impact of different patch sizes on prediction performance on the ETTm2 dataset. Colors correspond to different prediction horizons, and the black line represents the average MSE or MAE.}
  \label{error_vs_patchsize_ETTm2}
\end{figure}

\paragraph{Patch size} We conducted a study on the impact of different patch sizes (16, 32, 48, 56, 64) on prediction performance across three datasets: Weather, ETTh2, and ETTm2. The results are presented in Figs. \ref{error_vs_patchsize_Weather} to \ref{error_vs_patchsize_ETTm2}.

Different datasets and prediction horizons may favor different patch sizes. As shown in Fig. \ref{error_vs_patchsize_Weather}, on the Weather dataset, patch sizes of 32 and 64 achieved the best prediction performance, yielding the lowest MAE and MSE values. Similarly, Fig. \ref{error_vs_patchsize_ETTh2} demonstrates that a patch size of 32 leads to optimal results on the ETTh2 dataset. For the ETTm2 dataset, Fig. \ref{error_vs_patchsize_ETTm2} indicates that the best performance is obtained when the patch size ranges between 48 and 56.

These findings suggest that the optimal patch size is dataset-dependent and may also vary with the prediction horizon. Nevertheless, patch sizes of 32 or 64 generally provide robust predictive accuracy across different scenarios. Notably, since the ETTm2 dataset has a higher sampling frequency compared to ETTh2, it likely requires a larger patch size to effectively capture the finer-grained temporal dynamics and achieve superior forecasting performance.

\begin{figure}[!htbp]
  \centering 
 \includegraphics[width=1\textwidth]{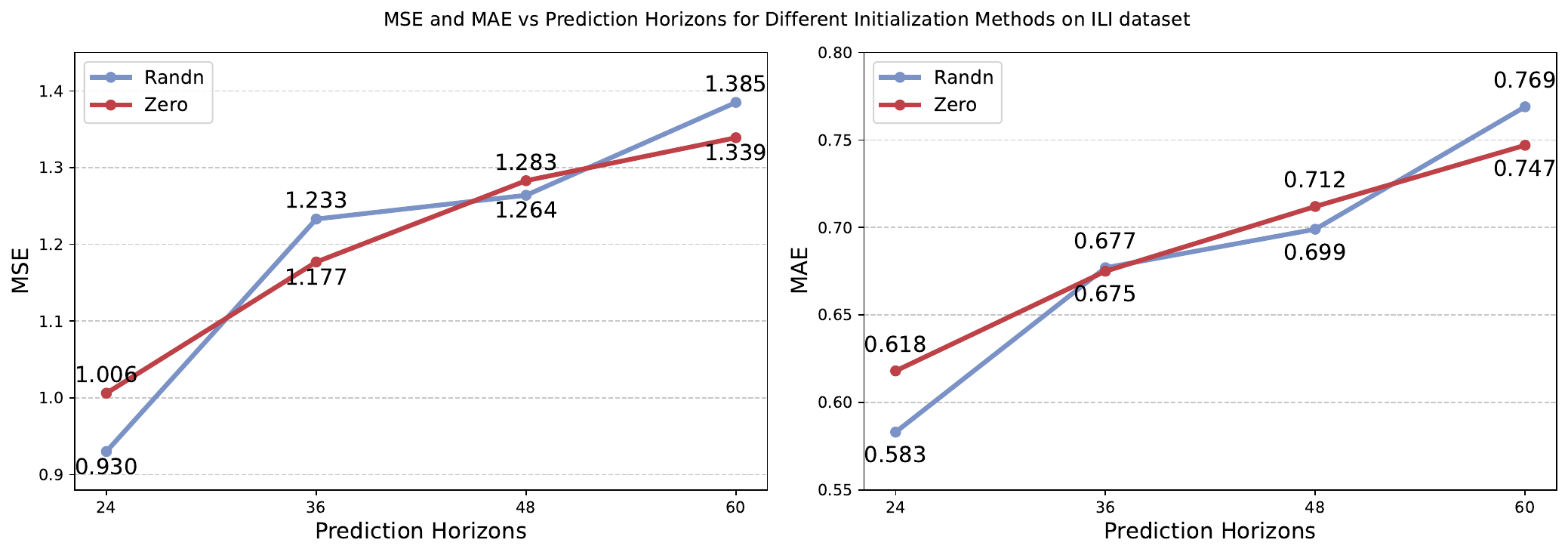} 
  \caption{Impact of different initialization methods for $\bm{z_0}$ on prediction performance under the ILI dataset.}
  \label{error_vs_init_ILI}
\end{figure}

\begin{figure}[!htbp]
  \centering 
\includegraphics[width=1\textwidth]{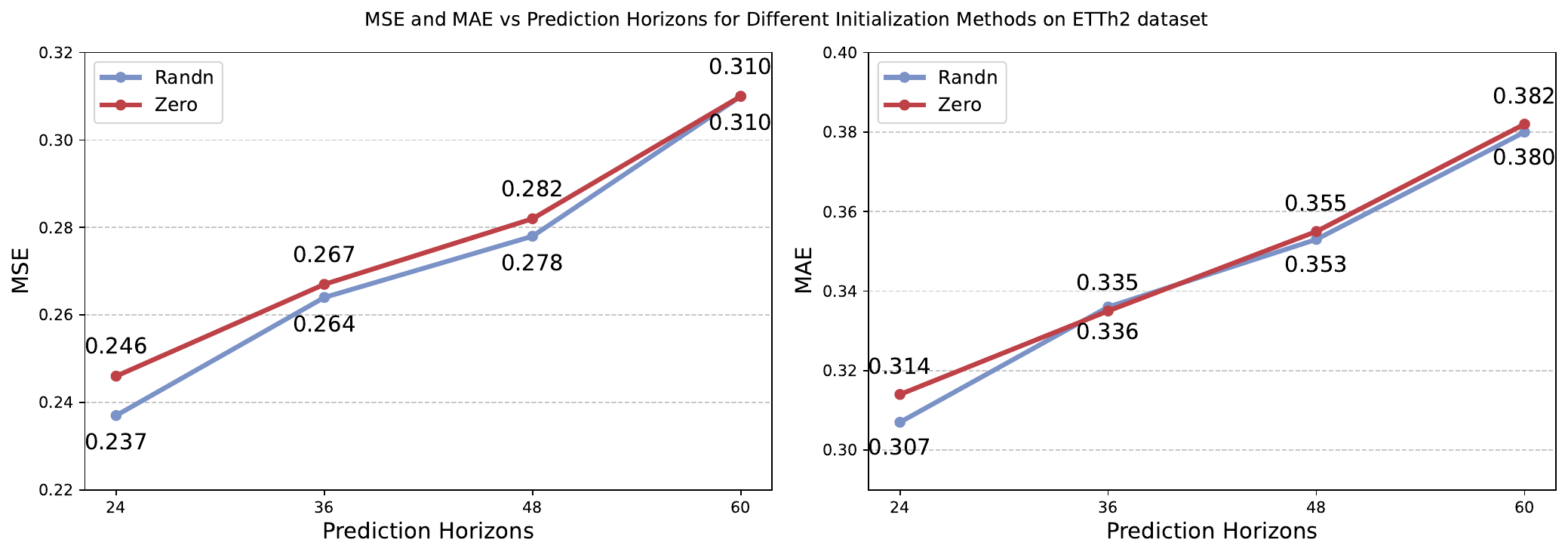} 
  \caption{Impact of different initialization methods for $\bm{z_0}$ on prediction performance under the ETTh2 dataset.}
  \label{error_vs_init_ETTh2}
\end{figure}

\paragraph{Initialization Method for $\bm{z_0}$} The initialization method for the initial latent state $\bm{z_0}$ in the SSM can also affect prediction performance. We compared two different initialization methods for $\bm{z_0}$—zero initialization (Zero) and Gaussian random initialization (Randn)—and evaluated their impact on prediction performance across different forecasting horizons on the ILI and ETTh2 datasets. The results are presented in Figs. \ref{error_vs_init_ILI} and \ref{error_vs_init_ETTh2}.

As illustrated in the figures, the choice of $\bm{z_0}$ initialization method leads to observable differences in prediction results. On the ILI dataset, Gaussian random initialization achieves an average MSE of 1.203 and an average MAE of 0.682 across different forecasting horizons, slightly outperforming zero initialization, which yields an average MSE of 1.201 and an average MAE of 0.688. Similarly, for the ETTh2 dataset, Gaussian random initialization attains an average MSE of 0.272 and an average MAE of 0.344, compared to zero initialization with an average MSE of 0.276 and an average MAE of 0.347.

Overall, Gaussian random initialization of $\bm{z_0}$ tends to yield better prediction performance. This may be because a randomly initialized latent state better captures the inherent stochasticity of the time series, thereby enhancing the model’s predictive accuracy and robustness.

\subsubsection{Representational Capacity of the Latent States.}

\begin{figure}[!htbp]
  \centering 
\includegraphics[width=0.9\textwidth, trim=12 20 12 18, clip]{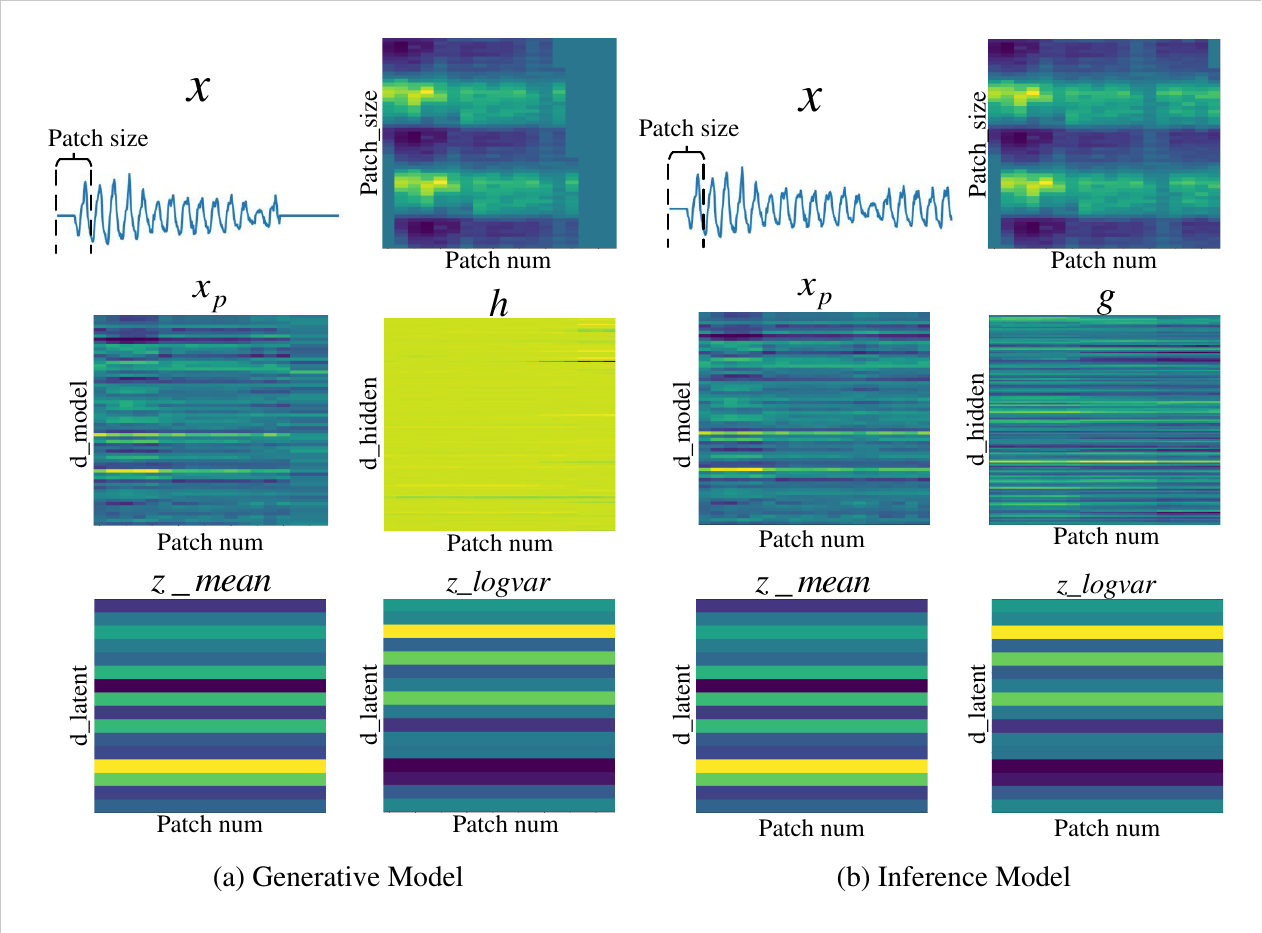} 
  \caption{Heatmaps of five tensor groups—\(x\), \(\bm{x_p}\), \(\bm{z}\), \(\bm{h}\), and \(\bm{g}\)—from StoxLSTM when predicting a specific instance in the Electricity dataset. Here, \(\bm{h}\) represents the output of the StoxLSTM recurrent unit during the generative process, while \(\bm{g}\) denotes the output of the bidirectional StoxLSTM recurrent unit during the inference process. Different colors correspond to different values, with brighter colors indicating larger magnitudes.}
  \label{fig:6}
\end{figure}

Fig. \ref{fig:6} shows heatmaps of four tensor groups—\( x_{1:L} \), \( \bm{x}_{p1:pN} \), \( \bm{z}_{1:N} \), and \( \bm{h}_{1:N} \)—extracted from a trained StoxLSTM model while forecasting a specific instance of the Electricity dataset. The four tensor groups correspond to the seasonal component obtained following the series decomposition block. Fig. \ref{fig:6}(a) presents heatmaps from the generative model, whereas Fig. \ref{fig:6}(b) shows those from the inference model.

Within the inference model, \( \bm{x}_{p1:pN} \) and \( \bm{g}_{1:N} \) vary little across patches but differ noticeably across dimensions within each patch. The latent variables capture this pattern well: both \( \bm{z\_mean}_{1:N} \) and \( \bm{z\_logvar}_{1:N} \) remain stable over patches while preserving significant variation across dimensions. This indicates that the latent representation encodes consistent temporal patterns while maintaining important feature-level distinctions.

Similarly, the generative model’s latent variables learned via variational inference exhibit near-uniformity across patches with meaningful dimension-wise variation. Such a well-structured latent representation, combined with the hidden states \( \bm{h}_{1:N} \), offers a comprehensive and informative summary of the time series, thereby supporting accurate and robust forecasting.

As demonstrated in Fig. \ref{fig:6}, the variational inference training converges effectively, and the latent states successfully capture the underlying temporal features of the data. This visualization validates both the model’s representational power and the interpretability of its latent space in modeling complex temporal dynamics.

\subsubsection{Model Efficiency}
\begin{figure}[!htbp]
    \centering
    \begin{subfigure}{0.48\textwidth}
        \centering
        \includegraphics[width=\textwidth]{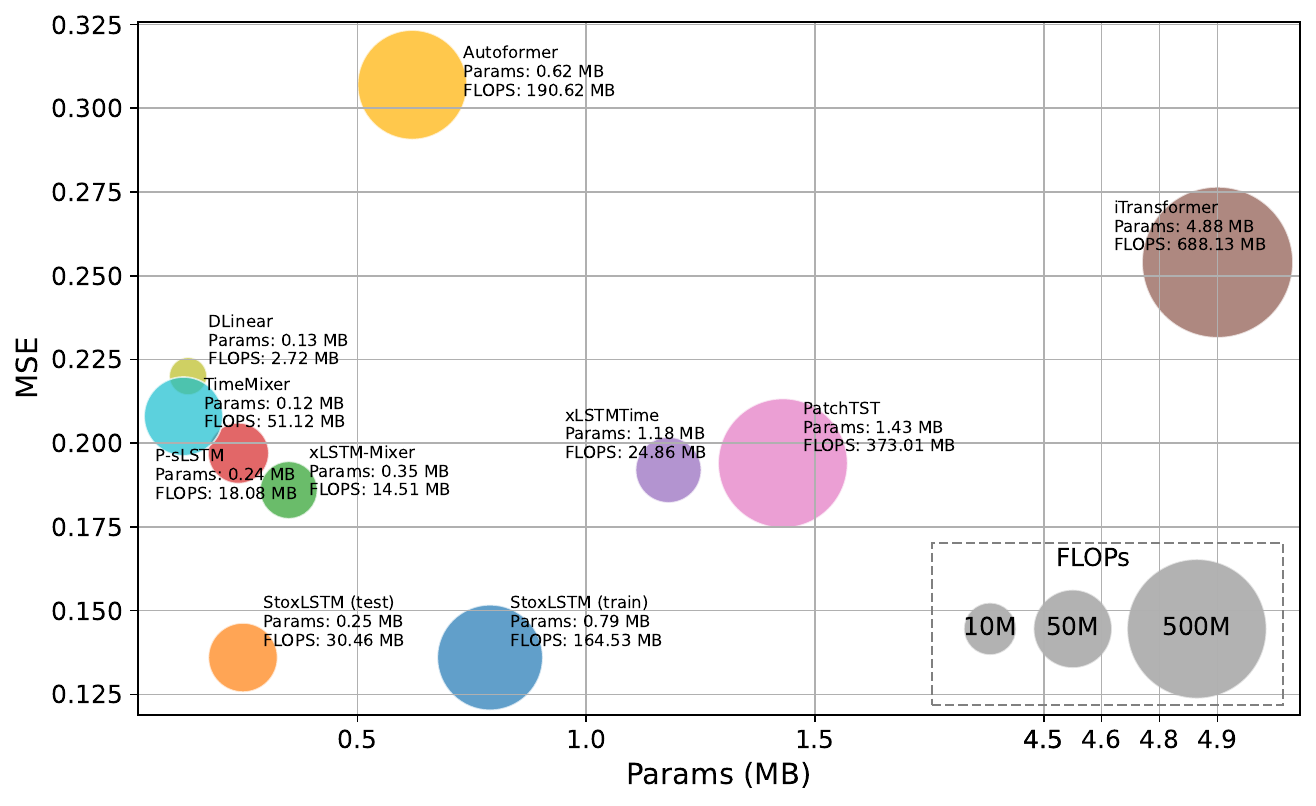}
        \caption{Weather Dataset}
        \label{fig:weather_efficiency}
    \end{subfigure}
    \hfill
    \begin{subfigure}{0.48\textwidth}
        \centering
        \includegraphics[width=\textwidth]{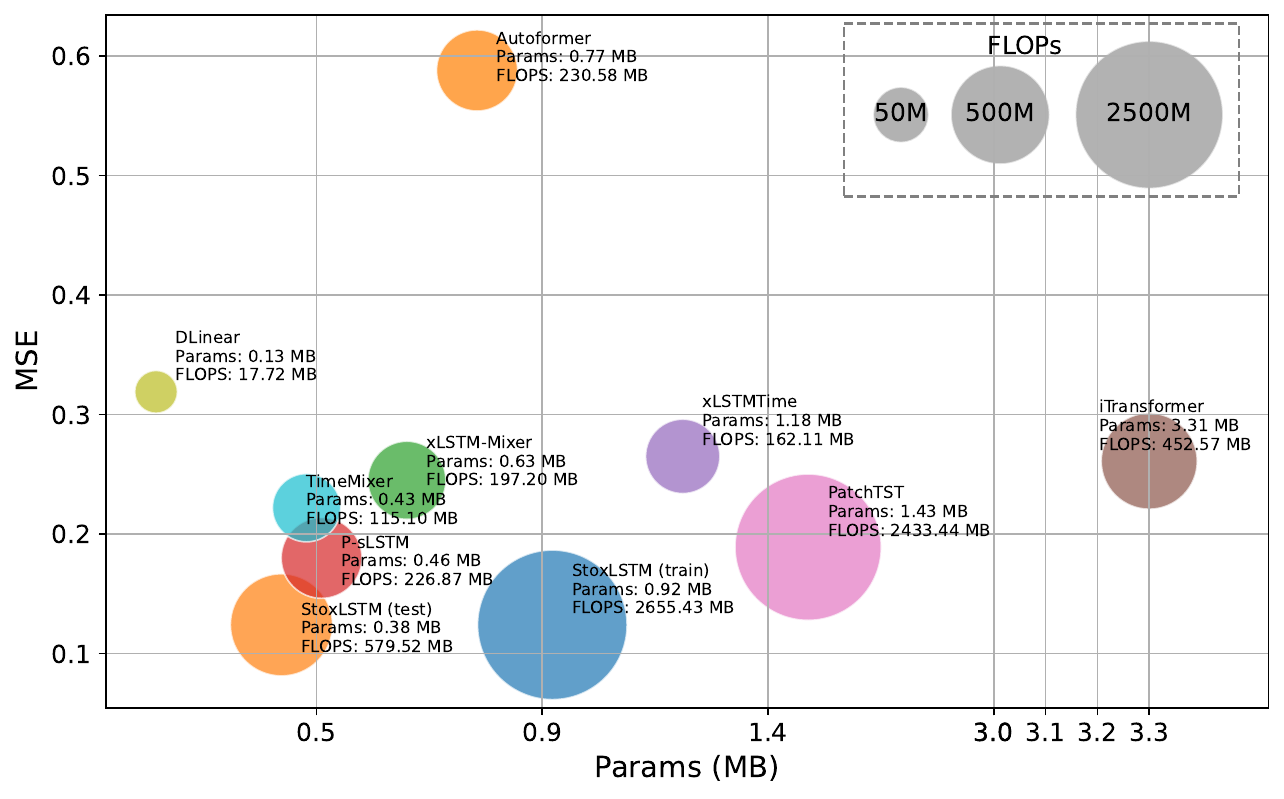}
        \caption{Solar Dataset}
        \label{fig:solar_efficiency}
    \end{subfigure}
    \caption{Comparison of model efficiency on Weather and Solar datasets. The prediction horizon for all models in both datasets is set to 192. Params denotes the number of model parameters, and FLOPs represents the theoretical floating-point operations required for a single forward pass.}
    \label{fig:model_efficiency_comparison}
\end{figure}

We compared the efficiency of different models on two datasets with varying dimensionality: the high-dimensional Solar dataset (137 dimensions) and the lower-dimensional Weather dataset (21 dimensions). The prediction horizon was set to 192 for both datasets. Efficiency metrics include MSE, the number of model parameters (Params), and the theoretical floating-point operations (FLOPs) required for a single forward pass. The results are presented in Fig. \ref{fig:model_efficiency_comparison}.

The results indicate that xLSTM-based models generally achieve lower MSE values, while their Params and FLOPs lie between those of Linear-based models and Transformer-based models. This demonstrates that xLSTM-based models offer a favorable balance between predictive accuracy and computational efficiency.

For StoxLSTM, the training phase requires an inference model to approximate the posterior distribution of the time series, resulting in substantially higher parameter counts and FLOPs during training. However, during testing, only the generative model is utilized, leading to a significant reduction in both model parameters and computational costs. Notably, StoxLSTM’s generative model achieves high efficiency, as its recurrent units enhance the original xLSTM structure without introducing additional complex modules. Future work will focus on simplifying the inference model to further enhance overall efficiency.

\section{Conclusion} \label{conclusion}
In this paper, we propose StoxLSTM, a stochastic xLSTM architecture. By integrating a designed SSM, StoxLSTM effectively captures the dynamic evolution of time series data. Our model demonstrates a strong capability in both long- and short-term forecasting tasks. Extensive experiments on multiple publicly available benchmark datasets show that StoxLSTM consistently outperforms existing state-of-the-art methods in terms of prediction accuracy and robustness. These results validate the effectiveness of combining the xLSTM framework with the SSM design to enhance temporal modeling.

However, we impose a Gaussian assumption on the latent variables, which may restrict the model’s ability to fully capture highly complex or multimodal temporal dynamics present in real-world time series. Second, due to the VAE-like deep state space model structure, StoxLSTM tends to produce relatively smooth predictions, limiting its capability to accurately model and forecast fine-grained short-term fluctuations and detailed temporal variations. Finally, the training process of StoxLSTM relies on an inference model to approximate the posterior distribution of latent variables, which significantly increases the number of parameters and computational overhead. Future work will focus on relaxing the Gaussian assumption, enhancing the model’s sensitivity to short-term dynamics, and developing more efficient inference mechanisms to reduce computational costs.

\section{Acknowledgment}
This work was supported in part by the National Natural Science Foundation of China under Grant 62301031, in part by the National Key Laboratory of Science and Technology on Space-Born Intelligent Information Processing under Grant TJ-01-25-01, and in part by the Young Elite Scientists Sponsorship Program by CAST under Grant YESS20240231.

\newpage
\appendix
\section{ELBO Derivation of StoxLSTM} \label{app A}
The KL divergence between the approximate posterior probability \({{q_\varphi }\left( {{z_{1:L + T}}|{x_{1:L + T}}} \right)}\) and the posterior probability \({p\left( {{z_{1:L + T}}|{x_{1:L + T}}} \right)}\) in StoxLSTM is derived as follows: 

\begin{align*}
& {\rm{KL}}\left( {q_\varphi \left( z_{1:L + T} | x_{1:L + T} \right) \| p \left( z_{1:L + T} | x_{1:L + T} \right)} \right) \\
= \, & \mathbb{E}_{q_\varphi \left( z_{1:L + T} | x_{1:L + T} \right)} \left[ \log q_\varphi \left( z_{1:L + T} | x_{1:L + T} \right) - \log p \left( z_{1:L + T} | x_{1:L + T} \right) \right] \\
= \, & \mathbb{E}_{q_\varphi \left( z_{1:L + T} | x_{1:L + T} \right)} \left[ \log q_\varphi \left( z_{1:L + T} | x_{1:L + T} \right) - \log p \left( z_{1:L + T} | x_{1:L + T}, x_{1:L} \right) \right] \\
= \, & \mathbb{E}_{q_\varphi \left( z_{1:L + T} | x_{1:L + T} \right)} \left[ \log q_\varphi \left( z_{1:L + T} | x_{1:L + T} \right) - \log p \left( x_{1:L + T}, z_{1:L + T} | x_{1:L} \right) + \log p \left( x_{1:L + T} | x_{1:L} \right) \right] \\
= \, & \mathbb{E}_{q_\varphi \left( z_{1:L + T} | x_{1:L + T} \right)} \left[ \log q_\varphi \left( z_{1:L + T} | x_{1:L + T} \right) - \log p \left( x_{1:L + T}, z_{1:L + T} | x_{1:L} \right) \right] + \log p \left( x_{1:L + T} | x_{1:L} \right)
\end{align*}

Then, \({{q_\varphi }\left( {{z_{1:L + T}}|{x_{1:L + T}}} \right)}\) and \({p\left( {{x_{1:L + T}},{z_{1:L + T}}|{x_{1:L}}} \right)}\) can be factorized over time according to Eq. \eqref{eq1} and \eqref{eq2}:

\begin{align*}
& \mathbb{E}_{q_\varphi(z_{1:L+T} | x_{1:L+T})} \left[ \log q_\varphi(z_{1:L+T} | x_{1:L+T}) - \log p(x_{1:L+T}, z_{1:L+T} | x_{1:L}) \right] + \log p(x_{1:L+T} | x_{1:L}) \\
= \, & \sum_{t=1}^{L+T} \mathbb{E}_{q_\varphi(z_{1:L+T} | x_{1:L+T})} \left[ \log q_\varphi(z_t | z_{1:t-1}, x_{1:L+T}) - \log p_\theta(z_t | x_{1:t-1}, z_{1:t-1}, x_{1:L}) - \log p_\theta(x_t | x_{1:t-1}, z_{1:t}, x_{1:L}) \right] \\
& \quad + \log p(x_{1:L+T} | x_{1:L}) \\
= \, & \sum_{t=1}^{L+T} \mathbb{E}_{q_\varphi(z_{1:L+T} | x_{1:L+T})} \left[ \log q_\varphi(z_t | z_{1:t-1}, x_{1:L+T}) - \log p_\theta(z_t | x_{1:t-1}, z_{1:t-1}, x_{1:L}) \right] \\
& \quad - \sum_{t=1}^{L+T} \mathbb{E}_{q_\varphi(z_{1:L+T} | x_{1:L+T})} \left[ \log p_\theta(x_t | x_{1:t-1}, z_{1:t}, x_{1:L}) \right] + \log p(x_{1:L+T} | x_{1:L}) \\
= \, & \sum_{t=1}^{L+T} \mathbb{E}_{q_\varphi(z_{t+1:L+T} | x_{1:L+T})} \Bigg[
    \mathbb{E}_{q_\varphi(z_{1:t-1} | x_{1:L+T})} \Bigg[
        \mathbb{E}_{q_\varphi(z_t | z_{1:t-1}, x_{1:L+T})} \Big[
            \log q_\varphi(z_t | z_{1:t-1}, x_{1:L+T}) - \log p_\theta(z_t | x_{1:t-1}, z_{1:t-1}, x_{1:L})
        \Big]
    \Bigg]
\Bigg] \\
& \quad - \sum_{t=1}^{L+T} \mathbb{E}_{q_\varphi(z_{t+1:L+T} | x_{1:L+T})} \Bigg[
    \mathbb{E}_{q_\varphi(z_{1:t-1} | x_{1:L+T})} \Bigg[
        \mathbb{E}_{q_\varphi(z_t | z_{1:t-1}, x_{1:L+T})} \Big[
            \log p_\theta(x_t | x_{1:t-1}, z_{1:t}, x_{1:L})
        \Big]
    \Bigg]
\Bigg] \\
& \quad + \log p(x_{1:L+T} | x_{1:L})
\end{align*}

Moreover, based on the following expression for the KL divergence:
\begin{align*}
\mathrm{KL} & \left( q_\varphi \left( z_t | z_{1:t-1}, x_{1:L + T} \right) \,\|\, p_\theta \left( z_t | x_{1:t-1}, z_{1:t-1}, x_{1:L} \right) \right) \\
= \, & \mathbb{E}_{q_\varphi \left( z_t | z_{1:t-1}, x_{1:L + T} \right)} \Big[
    \log q_\varphi \left( z_t | z_{1:t-1}, x_{1:L + T} \right) - \log p_\theta \left( z_t | x_{1:t-1}, z_{1:t-1}, x_{1:L} \right)
\Big]
\end{align*} The above expression can be simplified as follows:


\begin{align*}
& \sum_{t=1}^{L+T} \mathbb{E}_{q_\varphi \left( z_{t+1:L+T} | x_{1:L+T} \right)} \Bigg[
    \mathbb{E}_{q_\varphi \left( z_{1:t-1} | x_{1:L+T} \right)} \Bigg[
        \mathbb{E}_{q_\varphi \left( z_t | z_{1:t-1}, x_{1:L+T} \right)} \Big[
            \log q_\varphi \left( z_t | z_{1:t-1}, x_{1:L+T} \right) - \log p_\theta \left( z_t | x_{1:t-1}, z_{1:t-1}, x_{1:L} \right)
        \Big]
    \Bigg]
\Bigg] \\
& \quad - \sum_{t=1}^{L+T} \mathbb{E}_{q_\varphi \left( z_{t+1:L+T} | x_{1:L+T} \right)} \Bigg[
    \mathbb{E}_{q_\varphi \left( z_{1:t-1} | x_{1:L+T} \right)} \Bigg[
        \mathbb{E}_{q_\varphi \left( z_t | z_{1:t-1}, x_{1:L+T} \right)} \Big[
            \log p_\theta \left( x_t | x_{1:t-1}, z_{1:t}, x_{1:L} \right)
        \Big]
    \Bigg]
\Bigg] + \log p \left( x_{1:L+T} | x_{1:L} \right) \\
= \, & \sum_{t=1}^{L+T} \mathbb{E}_{q_\varphi \left( z_{t+1:L+T} | x_{1:L+T} \right)} \Bigg[
    \mathbb{E}_{q_\varphi \left( z_{1:t-1} | x_{1:L+T} \right)} \Big[
        \mathrm{KL} \left( q_\varphi \left( z_t | z_{1:t-1}, x_{1:L+T} \right) \,\middle\|\, p_\theta \left( z_t | x_{1:t-1}, z_{1:t-1}, x_{1:L} \right) \right)
    \Big]
\Bigg] \\
& \quad - \sum_{t=1}^{L+T} \mathbb{E}_{q_\varphi \left( z_{t+1:L+T} | x_{1:L+T} \right)} \Bigg[
    \mathbb{E}_{q_\varphi \left( z_{1:t-1} | x_{1:L+T} \right)} \Bigg[
        \mathbb{E}_{q_\varphi \left( z_t | z_{1:t-1}, x_{1:L+T} \right)} \Big[
            \log p_\theta \left( x_t | x_{1:t-1}, z_{1:t}, x_{1:L} \right)
        \Big]
    \Bigg]
\Bigg] + \log p \left( x_{1:L+T} | x_{1:L} \right) \\
= \, & \sum_{t=1}^{L+T} \mathbb{E}_{q_\varphi \left( z_{t+1:L+T} | x_{1:L+T} \right)} \Bigg[
    \mathbb{E}_{q_\varphi \left( z_{1:t-1} | x_{1:L+T} \right)} \Bigg[
        \mathrm{KL} \left( q_\varphi \left( z_t | z_{1:t-1}, x_{1:L+T} \right) \,\middle\|\, p_\theta \left( z_t | x_{1:t-1}, z_{1:t-1}, x_{1:L} \right) \right) \\
& \qquad \quad - \mathbb{E}_{q_\varphi \left( z_t | z_{1:t-1}, x_{1:L+T} \right)} \left[ \log p_\theta \left( x_t | x_{1:t-1}, z_{1:t}, x_{1:L} \right) \right]
    \Bigg]
\Bigg] + \log p \left( x_{1:L+T} | x_{1:L} \right)
\end{align*}

Rearranging the above equation:

\begin{align*}
\log p \left( x_{1:L+T} | x_{1:L} \right) = \, & 
{\rm KL} \left( q_\varphi \left( z_{1:L+T} | x_{1:L+T} \right) \,\middle\|\, p \left( z_{1:L+T} | x_{1:L+T} \right) \right) \\
& + \sum_{t=1}^{L+T} \mathbb{E}_{q_\varphi \left( z_{t+1:L+T} | x_{1:L+T} \right)} \Biggl[
    \mathbb{E}_{q_\varphi \left( z_{1:t-1} | x_{1:L+T} \right)} \Biggl[
        \mathbb{E}_{q_\varphi \left( z_t | z_{1:t-1}, x_{1:L+T} \right)} \Biggl[
            \log p_\theta \left( x_t | x_{1:t-1}, z_{1:t}, x_{1:L} \right)  \Biggr] \\
& \qquad \qquad \quad - \mathrm{KL} \left( q_\varphi \left( z_t | z_{1:t-1}, x_{1:L+T} \right) \,\middle\|\, p_\theta \left( z_t | x_{1:t-1}, z_{1:t-1}, x_{1:L} \right) \right)
    \Biggr]
\Biggr]
\end{align*}

Since \({\rm{KL}}\left( {{q_\varphi }\left( {{z_{1:L + T}}|{x_{1:L + T}}} \right)||p\left( {{z_{1:L + T}}|{x_{1:L + T}}} \right)} \right)\)  is non-negative, we can derive the ELBO as follows:

\begin{align*}
\log p \left( x_{1:L+T} | x_{1:L} \right) \geq L \left( \theta, \phi; x_{1:L+T} \right) = \sum_{t=1}^{L+T} \Bigg( & \mathbb{E}_{q_\varphi \left( z_t | z_{1:t-1}, x_{1:L+T} \right)} \left[ \log p_\theta \left( x_t | x_{1:t-1}, z_{1:t}, x_{1:L} \right) \right] \\
& - \mathrm{KL} \left( q_\varphi \left( z_t | z_{1:t-1}, x_{1:L+T} \right) \| p_\theta \left( z_t | x_{1:t-1}, z_{1:t-1}, x_{1:L} \right) \right) \Bigg)
\end{align*}

Finally, based on the assumptions in Eq. (1) and (2), the ELBO of StoxLSTM can be derived as:

\begin{align*}
\log p \left( x_{1:L+T} | x_{1:L} \right) & \geq L \left( \theta, \phi; x_{1:L+T} \right) = \\
 \, & \sum_{t=1}^L \Bigg(
    \mathbb{E}_{q_\varphi \left( z_t | z_{t-1}, x_{1:L+T} \right)} \left[ \log p_\theta \left( x_t | z_t, x_{1:t-1} \right) \right] 
    - \mathrm{KL} \left( q_\varphi \left( z_t | z_{t-1}, x_{1:L+T} \right) \| p_\theta \left( z_t | z_{t-1}, x_{1:t-1} \right) \right)
\Bigg) \\
& + \sum_{t=L+1}^{L+T} \Bigg(
    \mathbb{E}_{q_\varphi \left( z_t | z_{t-1}, x_{1:L+T} \right)} \left[ \log p_\theta \left( x_t | z_t, x_{1:L} \right) \right] 
    - \mathrm{KL} \left( q_\varphi \left( z_t | z_{t-1}, x_{1:L+T} \right) \| p_\theta \left( z_t | z_{t-1}, x_{1:L} \right) \right)
\Bigg)
\end{align*}

\bibliographystyle{IEEEtran} 
\bibliography{main}

\end{document}